\documentclass[10pt,twocolumn,letterpaper]{article}
\usepackage{iccv}
\usepackage{times}
\usepackage{epsfig}
\usepackage{graphicx}
\usepackage{amsmath}
\usepackage{amssymb}
\usepackage{amsthm}  
\usepackage{subcaption}
\newcommand{\etalit}{\textit{et al.}}
\usepackage{float}
\usepackage[flushleft]{threeparttable}
\usepackage{authblk}
\usepackage[toc,page]{appendix}
\usepackage{multirow,tabularx} 
\usepackage{ltablex}

\usepackage{pifont}%
\usepackage{multicol}
\usepackage[linesnumbered,ruled]{algorithm2e}
\renewcommand{\mathbf}[1]{\boldsymbol{#1}}
\usepackage{multirow,tabularx} 
\usepackage[dvipsnames]{xcolor}
\DeclareMathOperator*{\argmax}{arg\,max}

\newcommand{\cmark}{\ding{51}}%
\newcommand{\xmark}{\ding{55}}%
\usepackage[pagebackref=true,breaklinks=true,letterpaper=true,colorlinks,bookmarks=false]{hyperref}
\newcommand*\samethanks[1][\value{footnote}]{\footnotemark[#1]}
\iccvfinalcopy % *** Uncomment this line for the final submission

\def\iccvPaperID{947} % *** Enter the ICCV Paper ID here

\ificcvfinal\fi
\begin{document}

%%%%%%%%% TITLE
\title{BAOD: Budget-Aware Object Detection}

%\author{Alejandro Pardo\\
%\and
 \author[1]{Alejandro Pardo \thanks{denotes equal contribution}}
      \author[2]{Mengmeng Xu \samethanks}
      \author[2]{Ali Thabet}
      \author[1]{Pablo Arbel\'aez}
      \author[2]{Bernard Ghanem}% <-this % stops a space
    \affil[1]{Universidad de los Andes, Colombia}
    \affil[2]{King Abdullah University of Science and Technology (KAUST), Saudi Arabia}
\maketitle
%\thispagestyle{empty}

%%%%%%%%% ABSTRACT
\begin{abstract}
   We study the problem of object detection from a novel perspective in which annotation budget constraints are taken into consideration, appropriately coined Budget Aware Object Detection (BAOD). When provided with a fixed budget, we propose a strategy for building a diverse and informative dataset that can be used to optimally train a robust detector. We investigate both optimization and learning-based methods to sample which images to annotate and what type of annotation (strongly or weakly supervised) to annotate them with. We adopt a hybrid supervised learning framework to train the object detector from both these types of annotation. % (equivalently, image- and instance-level annotation). 
   We conduct a comprehensive empirical study showing that a handcrafted optimization method outperforms other selection techniques including random sampling, uncertainty sampling and active learning. By combining an optimal image/annotation selection scheme with hybrid supervised learning to solve the BAOD problem, % proposed budget-aware approach can 
   we show that one can achieve the performance of a strongly supervised detector on PASCAL-VOC 2007 while saving $12.8\%$ of its original annotation budget. Furthermore, when  $100\%$ of the budget is used, it surpasses this performance by $2.0$ mAP percentage points.
   %\B{be consistent: use only strongly and weakly supervised all throughout the paper; sometimes you use these terms and other times you use weakly and fully; be consistent; change text and figures accordingly}
\end{abstract}

%%%%%%%%% BODY TEXT
\section{Introduction} \label{introduction}
Object detection in images is a fundamental computer vision problem with applications in many tasks including face/pedestrian detection \cite{hyperface,bai2018finding,hu2017finding,papadopoulos2017extreme,du2017fused,wang2018pcn},  counting \cite{chattopadhyay2017counting, kang2018crowd}, and visual search \cite{collomosse2017sketching, mu2018towards}.
Building a successful object detector  encompasses three main dimensions: (1) \textbf{the image dataset} to be annotated for training the detector. A larger dataset allows for a more accurate detector, but the number of training images is limited by the annotation budget;
(2) \textbf{the annotation scheme} used to label the training images. One could annotate either image-level labels (the categories of the objects are known but their locations are unknown, denoted weakly supervised annotation) or instance-level labels (both categories and locations are known, denoted strongly supervised annotation) \cite{papadopoulos2017extreme,coco,konyushkova2018learning,papadopoulos2016we};
(3) \textbf{the detection model}. Most works on object detection fix the first two dimensions and only explore the third. In fact, they tend to focus on optimizing the detection model based on one or more training datasets that typically provide the same kind of annotations.
In this paper, we fix the detection model and investigate  solutions in the first two dimensions of the problem.
%agregar por qué importante, para qué sirve esto, puede ser cuando se cmabia de aplicacion y no hay dataset publicos disponibles. 

\begin{figure}[t]
  \centering
  \begin{subfigure}[b]{1.0\linewidth}
    \includegraphics[width=8cm]{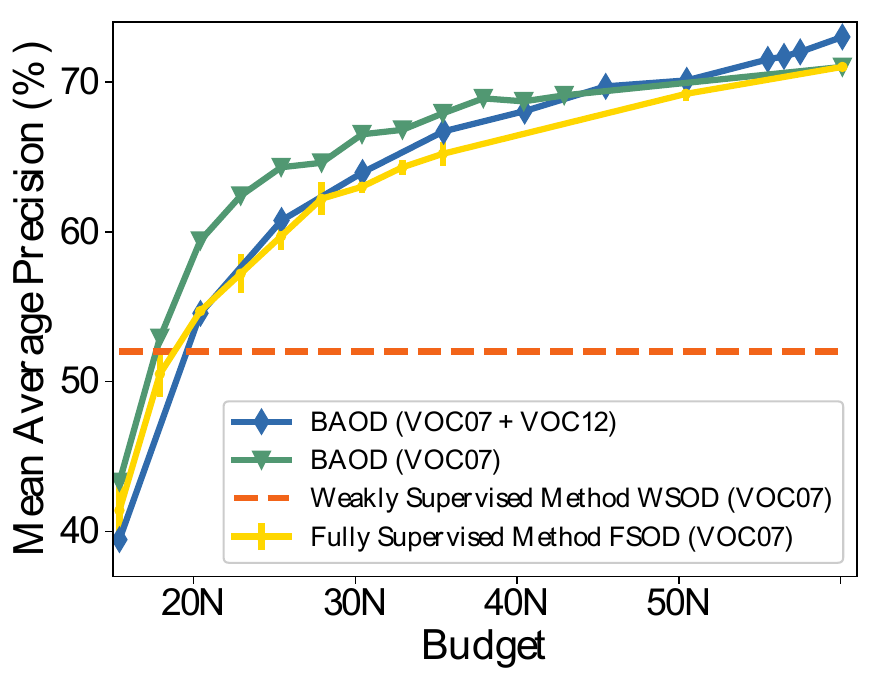}
  \end{subfigure}
  \caption{ \small
  \textbf{Budget-Aware performance of detectors with different levels of supervision.} The models are trained on PASCAL VOC 2007 trainval (VOC07), PASCAL VOC 2012 trainval (VOC12) as specified in the legend. Our proposed Budget aware object detection (BAOD) (green -$\nabla$- curve) has a higher mAP than FSOD (yellow -$\mid$- curve) and WSOD (orange -~- curve) methods at most budgets. Given a larger unlabeled image pool (Blue -$\diamondsuit$- curve, VOC07+VOC12), our BAOD can  reach a higher mAP using the same  budget needed to annotate  VOC07 with instance-level labels. Since the dataset is finite,  WSOD cannot increase its performance with more budget.
  %\B{make all caption font size small for both figures and tables}
  }
  \label{fig:main}
\end{figure}

A large group of object detectors fall under the umbrella of Fully-Supervised Object Detection (FSOD) \cite{yolov1, redmon2016yolo9000, rcnn, fast-rcnn, r-fcn, renNIPS15fasterrcnn, mask-rcnn, fpn}. It has been shown in recent years that these techniques can reach high detection performance, especially with the introduction of large datasets with strong annotations %(\ie instance-level annotations with object category and location) 
\cite{imagenet,pascal,coco}. This requirement makes FSOD methods expensive and time consuming. %For instance, the official protocol used to annotate ILSVRC ~\cite{imagenet} requires about 30 seconds per object box~\cite{AAAIW125350}.
In contrast, Weakly-Supervised Object Detection (WSOD) \cite{9,10,11,12,14,15,17,18,19,20,22} aims at building object detectors from cheaper but less informative image-level or weak annotations. %(\textit{i.e.} only object categories are given). 
% These weak annotations limit the detection performance of WSOD methods when compared to their FSOD counterparts trained on the same image dataset but with instance-level annotations. In fact, reaching FSOD performance is the gold standard for current WSOD methods. However, this comparison is inherently unfair, since WSOD training receives less information overall than FSOD, thus, registering a clear advantage for the latter.

In this paper, we propose a trade-off between dataset annotation cost and model precision in order to combine both weak and strong annotations and train an object detector with hybrid supervision. We put all the detectors on the same footing when different annotation schemes are available. Ideally, detectors should only be compared when they are trained using image datasets that offer the same amount of information (not necessarily the same number of images). Since this notion is difficult to define quantitatively, we take the \textbf{training budget} of a detector as a unifying surrogate measure. Here, we define budget as the effort, or cost, to annotate a dataset, thus combining the first two dimensions of the detection problem: dataset scale and annotation scheme. In fact, with a fixed budget and a set of unlabeled images, the number of images we can label depends highly on the annotation cost for each image. This cost varies significantly between image- and instance-level annotations. Typically, annotating a bounding box around an object in an image is significantly more expensive than simply annotating its category \cite{papadopoulos2017extreme, imagenet}. Therefore, an FSOD method with the same budget as a WSOD method contains fewer images in its training set. 

% Figure \ref{fig:main} fairly compares a FSOD \cite{renNIPS15fasterrcnn} and a WSOD method \cite{zhang2018w2f} at various budget levels, where the lowest budget corresponds to labelling the entire PASCAL VOC07 dataset with image-level annotations and the highest budget corresponds to the cost of labeling it with instance-level annotations. We observe that the WSOD method outperforms the FSOD one at low budgets. This result makes intuitive sense, since a low budget corresponds to many weakly annotated images and a much smaller number of strongly annotated ones, favoring WSOD. When the budget increases, the FSOD performance exceeds the one of the WSOD method. 

We explore strategies to build better object detector models when constrained with a training budget, a novel problem we coin \emph{budget-aware object detection} (BAOD). As such, we focus on choosing the best images for training and how to annotate them. For this, we survey several selection methods to sequentially choose both the image and type of annotation following an active learning paradigm. Additionally, we propose a novel hybrid training procedure for object detectors that can use both strong (instance-level) \emph{and} weak (image-level) annotations. %The combination of the active learning pipeline and the hybrid training is our Budget-Aware Object Detection (\textbf{BAOD}). 
Figure \ref{fig:main} shows that actively selecting images and their annotations to sequentially train a hybrid supervised detector outperforms its FSOD and WSOD counterparts, when the budget is larger than 20\% of the budget needed to annotate VOC2007 trainval at the instance-level. Moreover, the yellow curve shows that we can reach an even higher mAP if we use the same budget of VOC07 to annotate images from both datasets VOC07 and VOC12.
%In Pascal VOC2007 dataset\cite{pascal}, for example, we can save $12.8\%$ of budget to reach the same performance.

\paragraph{Contributions.}  \textbf{(1)} We propose the \textbf{BAOD} problem and present a new evaluation criterion \textbf{Budget-Average mAP} for object detection algorithms. This criterion takes into account both detection performance and budget. 
% Experiments in Section \ref{S:OPT} show that in many cases we can reach 95\% of state-of-the-art performance with only half of the total budget.
%\AKT{Link this to section parts of section 4. In section 4 put it in writing (final mAP is above 90\% of total)}
%\FCH{I felt that this contribution is not properly advertised in the intro. Maybe you need to mention somewhere how expensive is human annotation and you want to reduce it somehow -- something around the lines "human annotation is the opium of computer vision" lol.}
\textbf{(2)} We study several strategies (\eg optimization and learning-based) to select  both training images and their annotation scheme. %, including optimization and learning-based methods. 
%We compare different groups of solutions to the selection problem. The first group belongs to Optimization based methods in which we design the objective and the selection function and solve it with linear programming, results of these alternatives are shown in sections \ref{S:imageside} and \ref{S:OPT}. The second group belongs to learning based methods, mainly  Reinforcement Learning, in which we learn the objective function \ref{S:RL}. %\AKT{in RL section mention that this is the only method of this group that we tested}
\textbf{(3)} We propose a hybrid supervised learning strategy that combines category and localization information to train a robust detection model that handles both weak and strong annotations. 
%This framework inherits the advantages of both WSOD and FSOD  methods while avoiding their drawbacks. 
% To the best of our knowledge, we are the first to use hybrid training (\ie with both instance- and image-level annotations) in the context of object detection. 
% We show in Section \ref{S:HybridTraining} that our hybrid model beats fully supervised methods in common object detection metrics.
\textbf{(4)} Following a BAOD approach (\ie combining intelligent image and annotation scheme selection with hybrid supervised learning), we show that the mAP test performance on VOC07 can be improved by 2 percentage points for the same budget used to annotate the training set of VOC07 (by combining it with VOC12). %by combining both VOC07 and VOC12 sets and using only VOC07's budget. 
We also show the opposite, \ie that state-of-the-art test performance on VOC07 can be achieved, while saving $12.8\%$ of the budget used in strongly annotating its training set. 
% \textbf{(4)} By combining intelligent image and annotation scheme selection with hybrid supervised learning, we train a model that can outperform the state-of-the-art in two complementary ways. We improve performance on VOC07 by $2\%$ mAP combining both VOC07 and VOC12 train-val sets but using the same budget as the one used in VOC07. Also, we can achieve the current state-of-the-art performance on VOC07 \cite{jjfaster2rcnn}, while saving $12.8\%$ of its budget. 

\section{Related Work} \label{relatedwork}
\subsection{Fully Supervised Object Detection (FSOD)}
With the development of deep learning, many CNN based methods have been proposed to solve the FSOD problem, such as YOLO \cite{redmon2016yolo9000}, SSD \cite{liu2016ssd}, RCNN \cite{rcnn}, and its  variants \cite{renNIPS15fasterrcnn,fast-rcnn,r-fcn,mask-rcnn,fpn}. 
%Faster RCNN \cite{renNIPS15fasterrcnn} is a typical proposal based detection CNN, which balances both detection performance and computational efficiency. This method has become the \textit{de facto} framework for fully-supervised object detection. 
Although these methods achieve impressive detection results, providing them with large-scale instance-level (bounding-box) annotations for training is costly.

In this paper, we constrain the annotation budget and focus on how to reach the best detection performance within this budget by intelligently selecting between both weak and strong annotations. %reduce required annotations without  compromising the object detection performance. 
Su \etal \cite{AAAIW125350} reported that it takes $26$ seconds to draw one bounding-box without quality control, and $42$ seconds with it. There has been recent progress in developing tools to further reduce annotation time (most recently to $7$ seconds)~\cite{papadopoulos2016we,papadopoulos2017extreme,konyushkova2018learning}. Our work investigates a complementary aspect of annotation to emphasize that some images do not need to be strongly annotated to achieve excellent performance. While our discussion of budget aware object detection only considers the labor cost for image annotation, we note that sequentially training detection models also consume computational resources. However, given the continual development of hardware acceleration and re-organization, which leads to steady decrease in their cost, manual annotation remains the more expensive component in the detection problem.

\subsection{Weakly Supervised Object Detection (WSOD)}
If there are no instance-level labels available at training time, a WSOD method can be trained. Most classical approaches cast WSOD as a Multiple Instance Learning (MIL) problem \cite{16,17,18,19,9,22}. 
Bilen \etalit \cite{12} were the first to utilize a deep CNN (denoted as the weakly supervised deep detection network  or WSDDN) to solve this problem.  WSDDN selects positive samples by multiplying the score of recognition and detection. 
% Tang \etalit\cite{bai2017multiple} improved upon WSDDN by designing an online instance classifier refinement algorithm to alleviate the local-optimum problem that plagues WSDDN. 
Zhang \etalit\cite{zhang2018w2f} proposed a simple yet effective post-processing step to mine pseudo ground truth bounding boxes used for iterative FSOD training. % from WSOD scores. % excavation and pseudo ground truth adaptation algorithms. 
%In contrast to these methods, \cite{kumar2016track} mine the pseudo annotation from a previously trained tracker.

 We use weak supervision to improve pseudo object labels. These labels are generated from a previously trained detector, and are post-processed through a pseudo label mining process. The image-level labels help remove false predictions that correspond to a wrong object category.

\subsection{Hybrid Supervised Learning}
In our work, we study hybrid supervised learning for object detection. This type of learning exploits multiple types of supervision during training. Semi-supervised learning  is a special case of this family, since it learns a model from a set of labeled and unlabeled data.
Several works \cite{li2010optimol,rosenberg2005semi,chen2013neil,laine2016temporal, radosavovic2017data} try to solve the generic semi-supervised learning problem through  \textit{teacher-student learning}. They first train a \textit{teacher} model from the strongly annotated subset, then the predictions obtained from the teacher model on the unlabeled data are used to regress a second model that is called the student model.
% More generally, Hu \etal \cite{hu2018learning} proposed a novel weight transfer function to solve the instance segmentation problem from a hybrid dataset of instance- and  pixel- level annotations.
Ch\'eron~\etal \cite{cheron2018flexible} also used several kinds of annotations to train an activity recognition model, and they revealed that strongly annotating every training sample is not necessary to achieve noteworthy localization results in the video domain. 

In our case, since images can be labeled either weakly or strongly, a hybrid supervised dataset is always considered. % consists of image- and instance-level annotations. 
Inspired by Rosenberg's work \cite{radosavovic2017data}, we train a teacher- student model to use the hybrid dataset. Our teacher detector is learned from strong annotations only, while the student detector is learned from both ground-truth and/or processed pseudo object labels.

%\subsection{Sequentially Decision Making Process}
\subsection{Active Learning}
% Since we propose to sequentially select training images and their annotation type based on the current state of the detector subject to a budget constraint, we review some of the most related active learning literature here. 
In general, active learning is a sequential decision making process that iteratively selects the most useful examples an oracle should annotate and add to the labeled training set. It aims at training more accurate models with the minimum data required. This field has been widely studied in the context of image and video classification \cite{qi2008two,beluch2018power,konyushkova2017lal,kapoor2007active,zhu2017generative, hasan2015context},  object and action localization \cite{heilbron2018annotate, konyushkova2018learning, konyushkova2015introducing, papadopoulos2016we, kao2018localization}, human pose estimation \cite{ActiveHumanPose} and visual question answering \cite{misra2017learning}. A commonly used approach to selecting new training images is by means of their entropy score \cite{Wang2014ANA}. The intuition is that higher entropy examples attribute to more learning information. More recent research directly predicts the improvements of adding a new sample to the training set, and uses this measurement as a selection criteria \cite{konyushkova2017lal}.
% \F{One commonly used method is to calculate an entropy score of the training example \cite{Wang2014ANA} expecting that this score be directly related to the knowledge one can get from it. Another recent work proposed to predict the improvement if the sample is added to the training set\cite{konyushkova2017lal}.}
As specified in previous work, active learning aims at maximizing performance while minimizing the human cost in labeling the training samples \cite{qi2008two,ActiveHumanPose,heilbron2018annotate}.

The BAOD problem can also be formulated as a sequential decision making process. We study a few well-known selection techniques in our new active learning pipeline, which contains two active processes: (1) select the next batch of training images to annotate and (2) decide the type of annotation for each selected image. Thus, we focus on the annotation sequence that can provide the most useful information to the detector. To the best of our knowledge, we are the first to use active learning and hybrid training to study object detection.

\section{Budget Aware Object Detection (BAOD)} \label{approach}
\begin{figure}[!h]
\centering
\includegraphics[width=8cm]{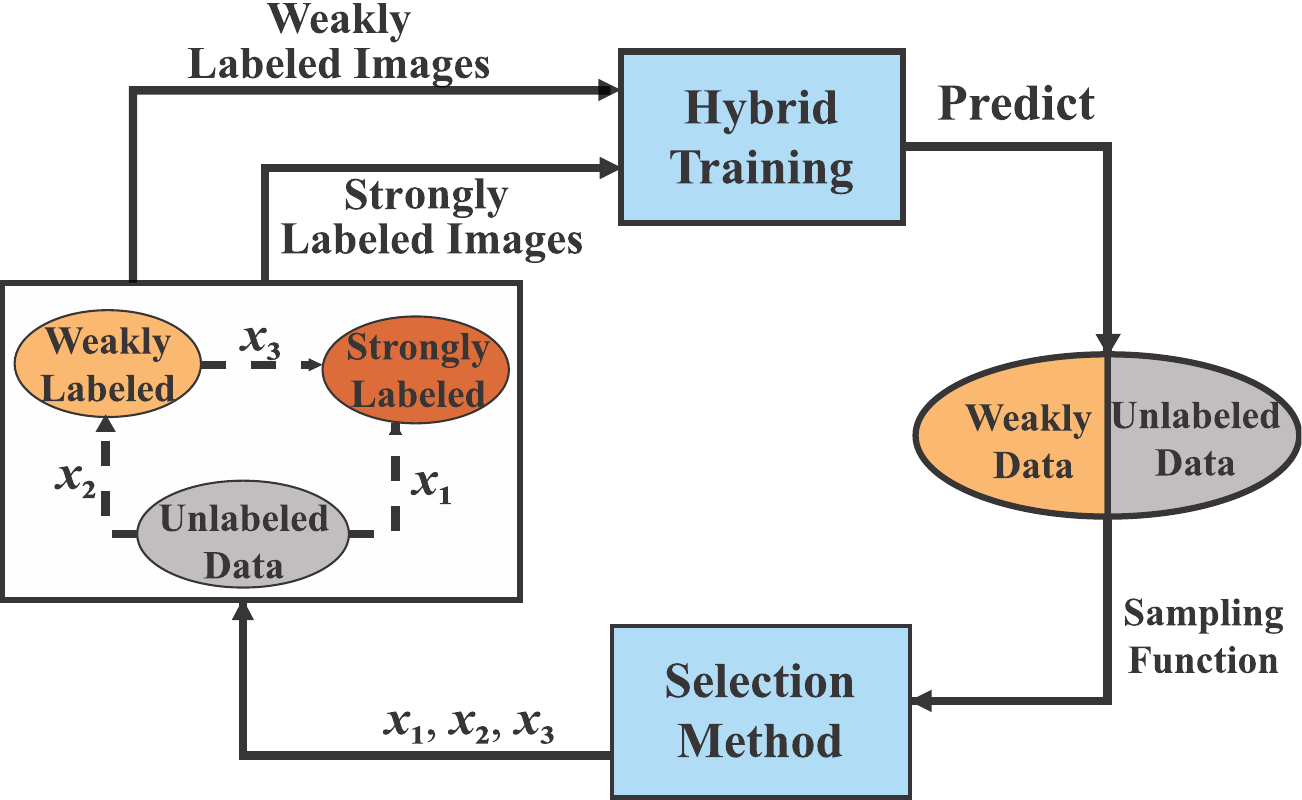}
\caption{\small \textbf{Overview of active learning pipeline to construct a hybrid labeled dataset.} For any weakly labeled or unlabeled image in the image pool (circular shapes), the selection method (bottom blue rectangle) decides which type of action to apply on the image based on the sample function and image status: weakly label ($x_1$) or strong label ($x_2,x_3$). Then such image is appended into the hybrid dataset and we train an object detection model with the hybrid supervision.
}
\label{fig:Map}
\end{figure}

In this section, we explore several ways to sequentially create an object detection dataset with a fixed budget, which includes both image-level and instance-level annotations. Then, we introduce a hybrid supervised learning procedure to take advantage of this hybrid labeled training set. A complete overview of the whole active learning procedure is shown in Figure \ref{fig:Map}.

\subsection{Hybrid Dataset using Active Learning} \label{activelearner}

Our hybrid supervised dataset should consider which images to include in training and how to label them. To tackle this problem, we propose an active hybrid learning framework which acts as an active learning agent to simulate the annotation process.
At every active step $\mathit{t}$, we have a budget constraint $\mathit{d}$ to spend on annotations. Each image in this set belongs to one of three pools: unlabeled $\mathit{U_t}$, weakly labeled $\mathit{W_t}$, or strongly labeled $\mathit{S_t}$. At each active step, the hybrid supervised dataset $\mathit{W_t}\cup\mathit{S_t}$ is used to train a hybrid detector, which will be described in detail in Section \ref{subsec: hybrid detection}. This detector is used to find a selection function that takes an action on $\mathit{U_t}$ or $\mathit{W_t}$ pools. As shown in Figure \ref{fig:Map}, we have three possible actions with their associated cost: (1) annotate strongly $\mathbf{x}_1$ (\textit{i.e.} send the image from $\mathit{U_t}$ to $\mathit{S_t}$), (2) annotate weakly $\mathbf{x}_2$ (\textit{i.e.} send the image from $\mathit{U_t}$ to $\mathit{W_t}$), or (3) strongly annotate a weakly annotated image $\mathbf{x}_3$ (\textit{i.e.} send the image from $\mathit{W_t}$ to $\mathit{S_t}$). Once the actions have been made, the image sets $\mathit{U_t}$,  $\mathit{W_t}$, and $\mathit{S_t}$ are updated. We proceed iteratively until we either run out of images in $\mathit{P_t}=\mathit{U_t \cup W_t}$ or run out of budget $\mathit{d}$.

To this end, we study three active learning sampling functions (Random, Uncertainty, and Learning Active Learning \cite{konyushkova2017lal}) within four action selection methods: Random Sampling (RS), Uncertainty Sampling (US), Optimization based on US and Optimization based on Learning Active Learning (LAL). 
%\B{no motivation of why these techniques were chosen; mention that they are well-known selection techniques in active learning literature} \F{Done}

For RS, we randomly choose an \textit{active batch} of images at each active step to include into $\mathit{W}_t$ or $\mathit{S}_t$. For US, images are sorted in descending order of uncertainty (measured by entropy) to choose high uncertainty images to train on with full supervision\footnote{We also conducted an experiment in which images in US were sorted in ascending order and low uncertainty images were chosen in every step. We discuss this in greater detail in the supplementary material.},
% \F{For US, images are sorted in descending order to choose high uncertainty images to train on with fully supervision\footnote{We also conducted an experiment in which images in US were sorted in ascending order and low uncertainty images were chosen in every step. We discuss this in greater detail in the supplementary material.}}, 
% \B{what is the motivation behind sorting in descending order? motivate why one would want high uncertainty score images to train on?} 
based on an uncertainty score denoted as $s_k$, and  the top images are included into $\mathit{W}_t$ or $\mathit{S}_t$. We adhere to the following annotation policy for RS and US when a budget constraint is enforced. For both RS and US, we prioritize weak labels first. In other words, so long as the budget constraint is not exceeded, (1) the image batch that is selected in each active step for these two methods will only contain images from $\mathit{U}_t$ that will be weakly labeled or, (2) if $\mathit{U}_t$ becomes empty, images from $\mathit{W}_t$ will be selected and get a strongly label. %\B{check for correctness; had to rewrite the paragraph}

%In both these selection methods, we prioritize $\mathit{W}_t$ In the first active steps, we weakly label images in $\mathit{U}_t$, then for the remaining active steps, we strongly label $\mathit{W}_t$. \B{there is no mention of budget at all here; this will confuse the reviewer}
%\B{how do we know whether to assign them to $\mathit{S_t}$ or $\mathit{W_t}$?}. 
There are several ways to evaluate  $s_k$ for an image \cite{Wang2014ANA,papadopoulos2017extreme}. We follow convention and model this score in Eq.~\ref{eq:US}:
\begin{equation}\label{eq:US}
{s_k =\frac{1}{M} \sum_{i=1}^M \sum_{p\in \mathbf{p}_i} - p\log(p)}
\end{equation} 
This is an entropy measure for each of the $M$ bounding boxes in an image using the classification score predicted by the current detection model. Collecting $M$ predictions from an image $I_k$, each prediction has a probability score vector $\mathbf{p}_i\in [0,1]^c$ for the $c$ object categories.

Below, we explain the remaining active selection methods: Optimization based on US and LAL.

\begin{figure*}[t]
\vspace{-10mm}
\centering
\includegraphics[width=17cm]{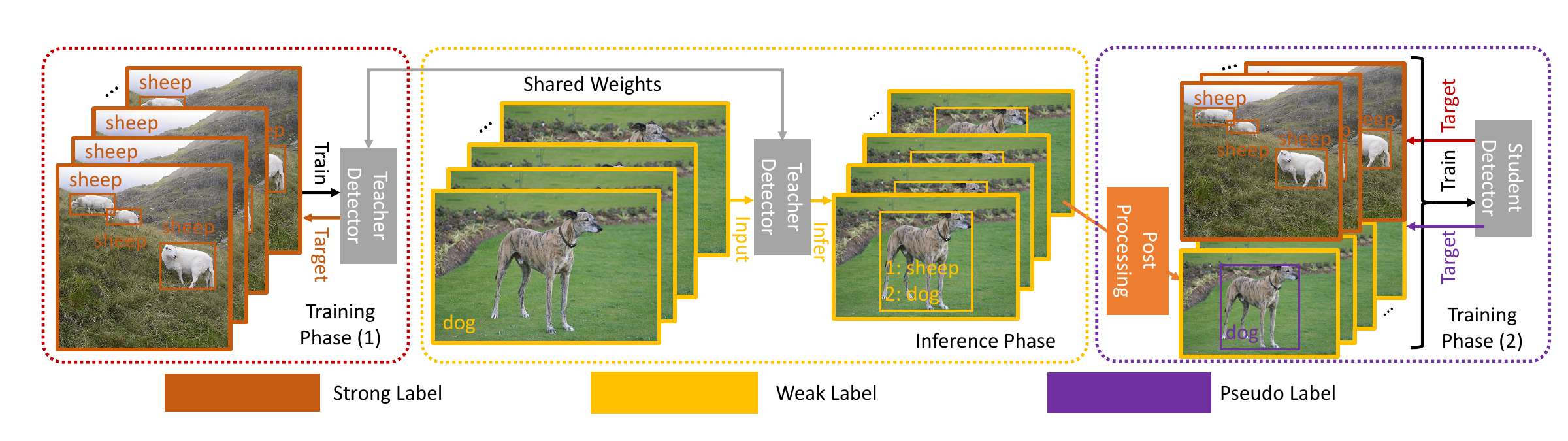}
\vspace{-5mm}
\caption{\small
\textbf{Illustration of the Hybrid Learning framework.} Given an image collection with hybrid labels, we firstly use a warm-up detection model to generate pseudo instance labels (\eg black solid rectangles in the inference phase.) 
After cleaning noise, the second detection model learns from both the ground truth and pseudo instance labels.
%\B{use the same font for text in the figure \eg sheep and Target}
% \B{come to my office to discuss details on how to improve this}
}
\label{fig:HybridLearning}
\end{figure*}

\subsubsection{Optimization Based Active Selection using US} \label{modelstate}
At the $t$-th active step, let $\mathit{P}_t=\{I_k\}_{k=1}^N$ be the $N$ images that can be further annotated, and $\mathbf{s}\in \mathbb{R}^N$ be their corresponding uncertainty scores (as defined above). 
We have three possible action vectors $\mathbf{x}_1, \mathbf{x}_2, \mathbf{x}_3 \in \{0,1\}^{N}$, as defined earlier. If the $k$-th element of $\mathbf{x}_i$ is $1$, we annotate the image $I_k$ using option $i$. Assume the next active batch of annotations has a linear impact on the model performance increment $\delta_t$. 
\begin{equation}
\delta_t = f_1(\mathit{P}_t)^{^\top}\mathbf{x}_1 + f_2(\mathit{P}_t)^{^\top}\mathbf{x}_2 + f_3(\mathit{P}_t)^{^\top}\mathbf{x}_3
\end{equation}
To quantitatively show the contribution of new images, we also assume that the uncertainty score is a complete statistic of an unlabeled or weakly labeled set of images. Then, we can simply approximate $(f_1, f_2, f_3)$ as linear   functions. % with input $\mathbf{s}$. 
We observe that many active learning studies \cite{USSampling,heilbron2018annotate} have experimentally shown that incorporating images with a higher uncertainty score in training may further improve the detector, but at the risk of having more difficulty in producing true positive predictions. Therefore, we model $(f_1, f_2, f_3)$ as linear functions that tend to favor actions $\mathbf{x_1}$, $\mathbf{x_3}$ over action $\mathbf{x_2}$ for images with high enough uncertainty score (\ie higher than the median). %To this end, we shift the uncertainty distribution to be balanced by subtracting its median value $\mu$. 
As such, and by combining the intuitions above, the expected increment in performance is modeled as:
\begin{equation}\label{eq:objective}
{\delta_t \approx \mathbf{s}^\top(\mathbf{x_1}+\mathbf{x_3})
+({\mu}\mathbf{1}- \mathbf{s})^\top\mathbf{x_2}}
\end{equation}

For the active selection method based on optimization using US, we seek to maximize $\delta_t$, while staying within the budget constraint. To model this latter constraint, we define  $a$ as the cost of strongly annotating an unlabeled image, $b$ as the cost of weakly annotating an unlabeled image, and $c$ as the cost of strongly annotating an already weakly labeled image. Therefore, this selection method seeks to solve the following binary optimization problem for $(\mathbf{x_1,x_2,x_3})$:
\begin{align}
\underset{\mathbf{x_1},\mathbf{x_2},\mathbf{x_3}\in \{0,1\}^{N}}{\max} ~\mathbf{s}^\top(\mathbf{x_1}+\mathbf{x_3}) 
+(\mu\mathbf{1}- \mathbf{s})^\top\mathbf{x_2}\notag\\
\text{s.t.}~~\begin{cases}
\mathbf{x_3}  \leq  \mathbf{\psi}\\
\mathbf{x_1}+\mathbf{x_2}  \leq  \mathbf{1}-\mathbf{\psi} \\
\mathbf{1}^\top (a  \mathbf{x_1}+b \mathbf{x_2}+c \mathbf{x_3}) \leq d \\
\mathbf{1}^\top (a  \mathbf{x_1}+b \mathbf{x_2}+c \mathbf{x_3}) \geq d-a
\end{cases}\label{eq:opt}
\end{align}

%%%%%%%%%%%%%%%%%%%%%%%%%%%%%
\iffalse
\begin{equation}\label{eq:opt1}
\begin{array}{rrclcl}
\displaystyle \max_{\mathbf{x_1},\mathbf{x_2},\mathbf{x_3}\in \{0,1\}^{N}} & \multicolumn{3}{l} 
{\mathbf{s}^\top(\mathbf{x_1}+\mathbf{x_3}) 
+(\mathbf{\mu}- \mathbf{s})^\top\mathbf{x_2}} \\
\textrm{s.t.} 
% & \mathbf{x_1}+\mathbf{x_2}+\mathbf{x_3}& \leq & \mathbf{1} \\
&\displaystyle \mathbf{x_3} & \leq & \mathbf{\psi} \\
&\displaystyle \mathbf{x_1}+\mathbf{x_2} & \leq & \mathbf{1}-\mathbf{\psi} \\
&\displaystyle \mathbf{1}^\top (a  \mathbf{x_1}+b \mathbf{x_2}+c \mathbf{x_3}) &\leq& d \\ &\displaystyle \mathbf{1}^\top (a  \mathbf{x_1}+b \mathbf{x_2}+c \mathbf{x_3}) &\geq& d-a \\
% &\displaystyle a \mathbf{x_1}+b \mathbf{x_2}+c \mathbf{x_3} &\leq& d
\end{array}
\end{equation}
\fi
%%%%%%%%%%%%%%%%%%%%%%%%%%%%%

Here, we define a vector $\psi$ as the indicator vector for images that have already been weakly annotated, \ie the $k$-th component of $\psi$ is 1 if $I_k\in\mathit{W}_t$ and 0 otherwise. Using this indicator, the first two constraints in Eq (\ref{eq:opt}) enforce that only one action is performed on each image, \ie among all the $k$-th components of $(\mathbf{x_1,x_2,x_3})$, only one can be 1. The third and fourth constraints enforce that the budget be used as much as possible in each active step.

%add constraints to make each of the action vectors consistent 
%Consider the constraints, we define $\psi$ to distinguish images' annotation state. $\psi_k=1$ only if the $k$-th image has already weakly annotated. For each solution, we have only one action for each image, and $\mathbf{x_3}$ (strongly annotate from weakly labeled data) can be positive only with a positive $\psi_k$.  In the budget constraints, if we use a budget $d$ in each active step, the cost, $\mathbf{1}^\top(a \mathbf{x}_1+b \mathbf{x}_2+c \mathbf{x}_3) $, in such active step should be in the range of $[d-a,d]$. Here $a,b,c$ are the cost to each labeling method, and $d$ is the given budget. % for each active step.

%In conclusion, we are going to solve the following constrained integer optimization problem to maximize the expected performance increment. The first two budget constraints force running out of the budget. The last three constraints make our solution consistent to actual operations.

The linear binary problem in Eq (\ref{eq:opt}) is NP-hard in general, so %, so it cannot be solved in polynomial time. 
exact solvers (\eg the Branch and Bound Algorithm) tend to have long run-times especially when $N$ is large. As a tradeoff between optimization accuracy and per active step run-time,  %takes more than 24 hours to find a global solution, 
we employ a conventional linear relaxation of the original problem to form an approximate linear program (LP), which can be efficiently solved at large-scale with off-the-shelf LP solvers. %its solution using a and uses a collection of linear restrictions. 
More  details about this problem are shown and proven in the supplementary material.

\subsubsection{Optimization Based Active Selection  using LAL} \label{LAL}
The uncertainty score $\mathbf{s}$  is not the only selection measure that has been studied in the active learning literature. In the Learning Active Learning (LAL) method proposed by Konyushkova \etal \cite{konyushkova2017lal}, the increment in performance function $\delta_t$ is learned to be  a function of the current model state as well. % of the absolute mAP  $\Delta$ as a learned target. 
More concretely, given images $I_k \in \mathit{P}_t$ and the current detection model, we build a feature vector $\mathbf{v}=[\mathbf{O}_t, \mathbf{s}]$ that concatenates both the current model state $\mathbf{O}_t$, represented as the average precision curves under five different Intersection over Union (IoU) thresholds, and the uncertainty scores $\mathbf{s}$. Following the LAL method in \cite{konyushkova2017lal}, we train a Support Vector Regression (SVR) model to regress the actual increment in mAP performance from $\mathbf{v}$ at each active step for both weak and strong annotation actions. Obviously, these SVRs are trained on a separate detection dataset than the one used to evaluate BAOD. At each active step and by denoting the output predictions of these SVRs as $\mathbf{h}_w$ and $\mathbf{h}_s$, we formulate the same constrained optimization problem in Eq (\ref{eq:opt}) but with the \emph{learned} objective:
\begin{equation}\label{eq:objectiveLAL}
{\delta_t \approx \mathbf{h}_w^\top \mathbf{x_2}
+\mathbf{h}_s^\top(\mathbf{x_1}+\mathbf{x_3})}
\end{equation}

\subsection{Hybrid Supervision for Object Detection} \label{subsec: hybrid detection}
At each active step, we train a hybrid supervised  detector using both strong and weak supervision, \ie using images from  $\mathit{W}_t$ and $\mathit{S}_t$ (refer to Figure \ref{fig:HybridLearning} for an illustration). 

\vspace{3pt}\noindent\textbf{Teacher-Student Model.} In this framework, we train an initial detector with the initial strongly annotated image set $\mathit{S}_0$. In each active step afterwards and to overcome undesirable local minima, we learn this detector using pretraining on Imagenet \cite{imagenet}. However, the model can also be fine-tuned from previous active steps to reduce computation time. Training on $\mathit{S}_t$ at each active step can only learn a decent detector, but it can also transfer knowledge to weakly annotated images. This detector works as a teacher that predicts objects in every image in the weakly labeled image set $\mathit{W}_t$. If these predictions are post-processed properly, they can be viewed as pseudo labels, which we merge with the ground truth instance-level labels in $\mathit{S}_t$ to train a fully supervised student detector. The student detector is also pretrained from Imagenet. 
Since it has both strong and weak annotations from training samples in $\mathit{W}_t \cup \mathit{S}_t$, we expect the student detector to perform better than its teacher. 

\vspace{3pt}\noindent\textbf{Post-processing.}
The predictions from the teacher model at each step have many redundant or erroneous pseudo labels. We use minimal knowledge to post-process them so as to focus our study on the active selection methods and the benefit of hybrid training. 
More concretely, given an image $I_k$ with a weak annotation $\mathbf{\omega} \in \{0,1\}^{c}$, where $c$ is the number of categories, we assume that the teacher model gives $M$ positive predictions with localization $\mathbf{P}\in \mathbb{R}^{M \times 4}$ (4 components defining a bounding box), classification $\mathbf{A}\in \{0,1\}^{M \times c}$, and the confidence score for each of the $M$ positive predictions $\mathbf{q}\in [0,1]^M$ . $\mathbf{P}$, $\mathbf{A}$ and $\mathbf{q}$ are obtained from the teacher model. %\AT{which object detector? you mean the teacher model?}.
%\B{this is never defined? is $m$ the image-level label for $I_k$ ?}. 
Among these $M$ predictions, we seek to mine the pseudo labels in the form of a sparse $M$-dimensional binary vector $\mathbf{y}$ that solves the following constrained optimization:
%Eventually, the pseudo label mining returns a sparse $M$-dimensional binary vector $\mathbf{y}$ as Eq.\ref{eq:PLM}, where $(\alpha=0.3,\beta=3)$.
\begin{align}
&\underset{\mathbf{y}\in \{0,1\}^{M}}{\max} ~\mathbf{y}^\top \mathbf{q}\notag\\
&\text{s.t.}~~\begin{cases}
\mathbf{y}^\top \mathbf{A} (\mathbf{1}-\mathbf{w}) =  0\\
\text{IoU}(\mathbf{P}_i, \mathbf{P}_j) \leq\alpha~~~ \forall~ y_i,y_j=1\\
1\leq\|\mathbf{y}\|_0 \leq \beta
\end{cases}\label{eq:PLM}
\end{align}

\iffalse
\begin{equation} \label{eq:PLM1}
\begin{array}{rrclcl}
\displaystyle \max_{\mathbf{y}\in \{0,1\}^{M}} & \multicolumn{3}{c}{\mathbf{y}^\top \mathbf{p}_m} \\
\textrm{s.t.} & \mathbf{y}^\top A (\mathbf{1}-\mathbf{w})& = & 0 \\
& \forall_{{y}_i,{y}_j=1} IoU(P_i, P_j) &\leq&\alpha\\
&\displaystyle |\mathbf{y}|_0 & \in & [1,\beta] 
\end{array}
\end{equation}
\fi 

The intuition behind solving this problem is that we seek to maximize the confidence score across all pseudo labels  indexed by $\mathbf{y}$. The first constraint enforces image-level consistency of the pseudo labels with the ground truth weak annotations. The second constraint removes predictions that are highly overlapping (similar to an NMS post-processing step). The third constraint enforces a sparsity condition on $\mathbf{y}$, thus limiting the number of possible pseudo labeled bounding boxes (potential objects) to be between 1 and $\beta$. In our experiments, we take $(\alpha=0.3,\beta=3)$.  Details on this optimization are in the supplementary material.

\section{Experiments and Analysis} \label{experiments}
\subsection{Experimental Setup}
\textbf{Datasets.} As in most WSOD methods, we use the PASCAL VOC 2007 (VOC07) or 2012 (VOC12) datasets \cite{pascal} to perform most of the experiments. Given the active learning pipeline of our method, we emulate the active learning procedure using VOC07-12 annotations to selectively annotate 20 categories in 5011 images in VOC07 or 16551 images in the union of VOC07 and VOC12 (VOC0712). %\B{emulate the oracle? what oracle? use simple words that make straightforward sense; rewrite this sentence}. 
All detection models are evaluated on the VOC07 test dataset. For each annotation type (weak or strong), we assume that each image has a fixed annotating cost/time, which is not necessarily true in practice but it simplifies the analysis. In most of the experiments, we set $(a=34.5,b=1.6,c=a-b)$, in unit seconds, according to the annotation procedure of \cite{imagenet}. In Section \ref{subsec: robustness to cost}, we vary these cost value to $(a=7,b=1.6,c=a-b)$ following the more efficient annotation procedure of \cite{papadopoulos2017extreme}\footnote{We left more discussion on the annotating time assumption in the supplementary material.}. %of strong label and weak label to be mostly (34.5,1.6) \cite{imagenet} in the unit of seconds, but we also implement experiments with (21,1.6) or (7,1.6) \cite{papadopoulos2017extreme} to show the robustness. 
The cost of fully annotating VOC07 trainval is denoted as $100\%$ (or total) budget for every experiment. 

\textbf{Evaluation Metrics.} To evaluate the active selection methods with the hybrid detector, we compute a budget-performance curve at various budget limits. The budget axis varies the percentage of the total budget, and the model detection performance is taken to be mAP. In doing so, we propose a new budget-aware metric denoted as \textbf{Budget-Average mAP}, measured as the normalized area under the budget-mAP curve for a certain budget range. We take three ranges $[10\%,30\%]$, $[30\%,50\%]$, and $[50\%,100\%]$ to evaluate our experiment such that three to five data points are located in each range and the metric is less affected by noise. The first range starts from $10\%$ since  we need a small fully labeled warm up set to initialize the fully supervised detector, and start our pipeline. This warm up set is randomly selected and fixed in all experiments. The last budget range is wider because the performance curve saturates at high budgets, and we observe more subtle changes in performance.
% \B{justify in one sentence why there are three ranges}.
All  budget-performance curves are shown in the supplementary material.

\textbf{Implementation Details.} In every active step, we choose Faster-RCNN as the object detection model (teacher and student), which is trained in the hybrid supervised way  described in Section \ref{subsec: hybrid detection}. %, and develops the framework on top of it. 
%Pablo suggest to remove it, because it can make the reviewers ask for more detectors:
%However, this model can be replaced by any \textit{off-the-shelf} object detector. 
Both teacher and student models use VGG16 as the backbone network pre-trained on ImageNet \cite{imagenet}. We follow the default setup in  \cite{jjfaster2rcnn} to train Faster-RCNN. During training, the total number of epochs is set to $10$, the learning rate is $0.01$ for the first $8$ epochs and $0.001$ for the remaining. The batchsize is set to 16 on four-GPU cluster nodes equipped with Titan Xp. The student model is cloned from the ImageNet pretrained VGG16 in every active step and has the same training schedule as the teacher model\footnote{The source code of the framework will be made publicly available. }. For LAL experiment, we collect 10 categories  from the MS-COCO dataset \cite{coco} to train the SVRs. Since the LAL method is dataset agnostic \cite{konyushkova2017lal}, we take the SVR training  categories to be different from the 20 categories in PASCAL VOC.

\subsection{Advantages of Uncertainty Sampling and Hybrid Training} \label{S:imageside}

% \begin{figure}[!h]
%   \centering
%     \includegraphics[width=0.9\linewidth]{FIGURE/Random.pdf}
%     % \includegraphics[width=\linewidth,height=6.5cm, keepaspectratio]{FIGURE/Random.pdf}
%   \caption{\small{\textbf{Budget-aware performance using fully and hybrid training pipelines with random and uncertainty selection.} Uncertainty  sampling ({orange}) is always better than random sampling {blue} selection, and hybrid training ({dark color}) is always better than FSOD ({light color}).} }
%   %\B{mAP not mAPs}
%   \label{fig:random}
% \end{figure}

\textbf{Uncertainty Sampling.} In order to explore the influence of the uncertainty score on model performance, we emulate the annotation process of the oracle using VOC07 and incorporate it into the active learning pipeline with \emph{only} strong annotations and use a FSOD method to train our model.
% \F{follow the  Fully Supervised Object Detection (FSOD) method to train the model}.
We use two selection methods (RS and US) to build the dataset for FSOD.
From the first two columns in Table \ref{tb:EXP1} that compare these two sampling methods, we see that collecting images with large uncertainty score is more effective and preferable than random sampling in the three budget ranges.

\textbf{Hybrid Training.} Table \ref{tb:EXP1} also shows the influence of the active selection scheme on model performance but when the hybrid supervised learning process is used to combine weak and strong annotations at each active step.
We again use RS and US to select the images in each step for hybrid training. It is clear that both hybrid methods outperform their corresponding FSOD counterpart at every budget range, especially when the budget range is low (under $50\%$). 

\begin{table}[h]
\caption{\small{\textbf{Budget-Average mAP using fully and hybrid training pipelines with random and uncertainty selection.} Uncertainty  sampling is always better than random sampling selection, and hybrid training is always better than FSOD.}}
\vspace{-5pt}
\begin{center}
\small
\label{tb:EXP1}
\begin{tabular}{c||c|c|c|c}

\hline
\textbf{Selection Method} & \multicolumn{2}{c}{\textbf{FSOD}} & \multicolumn{2}{c}{\textbf{Hybrid} }\\  \hline 
\textbf{Sampling Function} & \textbf{RS} & \textbf{US} & \textbf{RS} & \textbf{US} \\ \hline \hline
\textbf{Low Budget Range}    & 52.5  & 53.1 & \textbf{56.4} & 55.1 \\ \hline
\textbf{Mid Budget Range}    & 62.5  & 63.6 & 64.5 & \textbf{65.8} \\ \hline
\textbf{High Budget Range}   & 67.9  & 68.7 & 68.7 & \textbf{69.3} \\ \hline
\end{tabular}
\end{center}
\vspace{-10pt}
\end{table}

\subsection{Optimization-Based Active Selection}\label{S:OPT}

% \begin{figure}[h!]
%   \centering
%     \includegraphics[width=0.9\linewidth]{FIGURE/Opt.pdf}
%     % \includegraphics[width=\linewidth,height=6.5cm, keepaspectratio]{FIGURE/Opt.pdf}
%   \caption{\small{\textbf{\F{Budget-aware performance using 
%   % \B{where is FSOD? more evidence that the paper was not read completely}
%   % \B{clearly state that hybrid supervised learning is used here for all methods} 
%   simple hybrid training and optimization methods.}} US based optimization (yellow) is slightly better than LAL one (gray). The optimization methods perform better than the simple hybrid random selection (orange) and uncertainty selection (blue bars) methods in the three budget ranges.}}
%   \label{fig:Hybrids}
% \end{figure}

\begin{figure*}[!ht]
\centering
\includegraphics[width=0.34\linewidth]{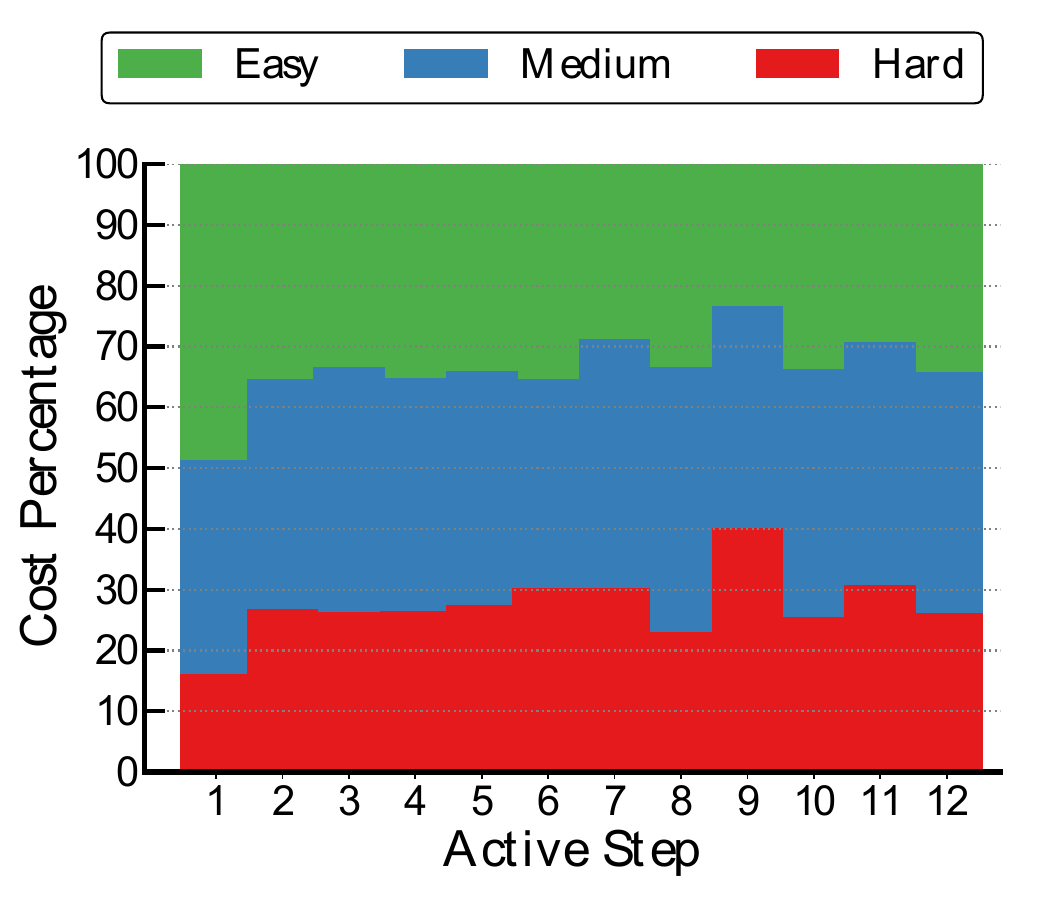}
\includegraphics[width=0.64\linewidth]{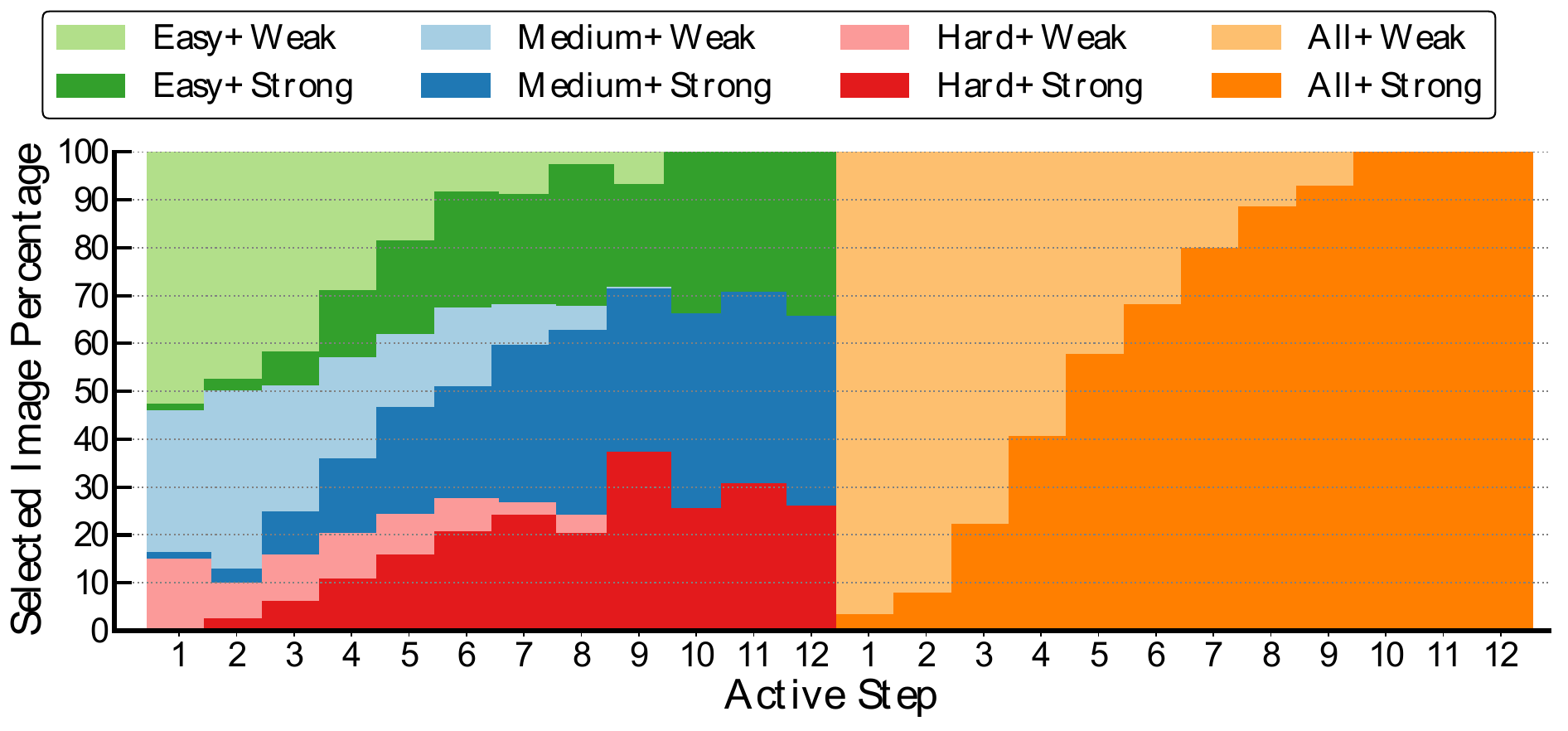}
\caption{\small{
\textbf{Comparison of the cost and image number on every active selection step.} 
\textit{Left}: {Budget usage distribution to learn different difficulty categories.} More budget is used to annotate Easy images (green area) at beginning. The cost spent on Hard images (red area) grows up when the active model is mature. 
\textit{Right}: {Selected images distribution in different difficulty categories and different annotation type.} The selection agent gives more weak annotations (light color) at the first steps. Given more budget, the proportion of strong annotations (dark color) increase. We run out of unlabeled images after 9-th step. The mapping is motivated and shown in the supplementary material.
} }
% \B{remove titles}
\label{fig:analyse}
\end{figure*}

Here, we evaluate the optimization-based selection methods (using US and LAL) described in Section \ref{activelearner}, when they are combined with hybrid supervised training. Table \ref{tb:EXP2} compares these methods by measuring their average mAP at each budget range and compares them to the RS and US selection methods for references. These results indicate that the optimization methods are the best ways to combine and use the hybrid annotations at every budget, where the optimization-based US method is slightly better than its LAL counterpart. As such, we denote the former method as the BAOD approach, which was mentioned and  highlighted in Figure \ref{fig:main}. Interestingly, we observe that the RS method requires $62\%$ of the total VOC07 budget (all images are strongly annotated) to achieve $95\%$ of the detection performance at that budget (\ie $67.4\%$ mAP). In comparison, the BAOD method requires only $48.5\%$ of the total budget to reach the same performance. This performance gap attests to the effectiveness of this method. We include more results in the supplementary material.
 
\begin{table}[h]
\caption{\small{\textbf{Budget-Average mAP using simple hybrid training and optimization methods.} US based optimization is slightly better than LAL one. The optimization methods perform better than the simple hybrid random selection and uncertainty selection methods in the three budget ranges.}}
\begin{center}
\small
\label{tb:EXP2}
\begin{tabular}{c||c|c|c|c}
\hline
\textbf{Selection Method} & \multicolumn{2}{c}{\textbf{Hybrid}} & \multicolumn{2}{c}{\textbf{Optimization} }\\  \hline 
\textbf{Sampling Function} & \textbf{RS} & \textbf{US} & \textbf{LAL} & \textbf{US} (BAOD) \\ \hline \hline
\textbf{Low Budget Range}    & 56.4  & 55.1 & 56.3 & \textbf{57.1} \\ \hline
\textbf{Mid Budget Range}    & 64.5  & 65.8 & 65.9 & \textbf{66.0} \\ \hline
\textbf{High Budget Range}   & 68.7  & 69.3 & 69.3 & \textbf{69.5} \\ \hline
\end{tabular}
\end{center}
\vspace{-10pt}
\end{table}

\subsection{Effect of Per-Image Annotation Cost} \label{subsec: robustness to cost}

The cost of strong annotations can vary due the  annotation strategy that is used. For instance, Papadopoulos \etal~\cite{papadopoulos2017extreme} created a method that reduces the time needed to draw bounding boxes to 7 seconds per image on average. Here, we use this cost ($a=7$) to study the behavior of our BAOD method given a smaller gap between strong and weak annotation. We report these results in Table \ref{tb:EXP3}, which depicts the best performing FSOD method and the RS hybrid baseline for reference. We observe the same relative behavior between the methods as in the case when $a$ was several times higher. However, the performance gap between these methods decreases in this case because the number of images that can be weakly annotated for the cost of one strong annotation is much smaller.

\begin{table}[h]
\caption{\small{\textbf{Budget-Average mAP using a lower cost for strong annotations.} If we assume the weak and strong annotation costs are more close (7 seconds and 1.5 seconds), US based optimization (BAOD) still performs better than the simple hybrid random selection  and uncertainty selection methods in the three budget ranges.} }
\vspace{-9pt}
\begin{center}
\label{tb:EXP3}
\small
\begin{tabular}{c||c|c|c}
\hline
\textbf{Selection Method} & \textbf{FSOD} & \textbf{Hybrid} & \textbf{Optimization} \\ \hline 
\textbf{Sampling Function} & \textbf{RS} & \textbf{US} & \textbf{US} (BAOD) \\ \hline 
\hline
\textbf{Low Budget Range}    & 53.1  & 52.3 & \textbf{54.2} \\ \hline
\textbf{Mid Budget Range}    & 62.8  & 63.1 & \textbf{63.6} \\ \hline
\textbf{High Budget Range}   & 68.3  & 68.6 & \textbf{69.3} \\ \hline
\end{tabular}
\end{center}
\end{table}
\vspace{-9pt}

\subsection{Improving Detection using Fixed Budget }
\label{S:APP}

\begin{figure*}[!h]
\vspace{-20pt}
\centering
\includegraphics[width=\linewidth]{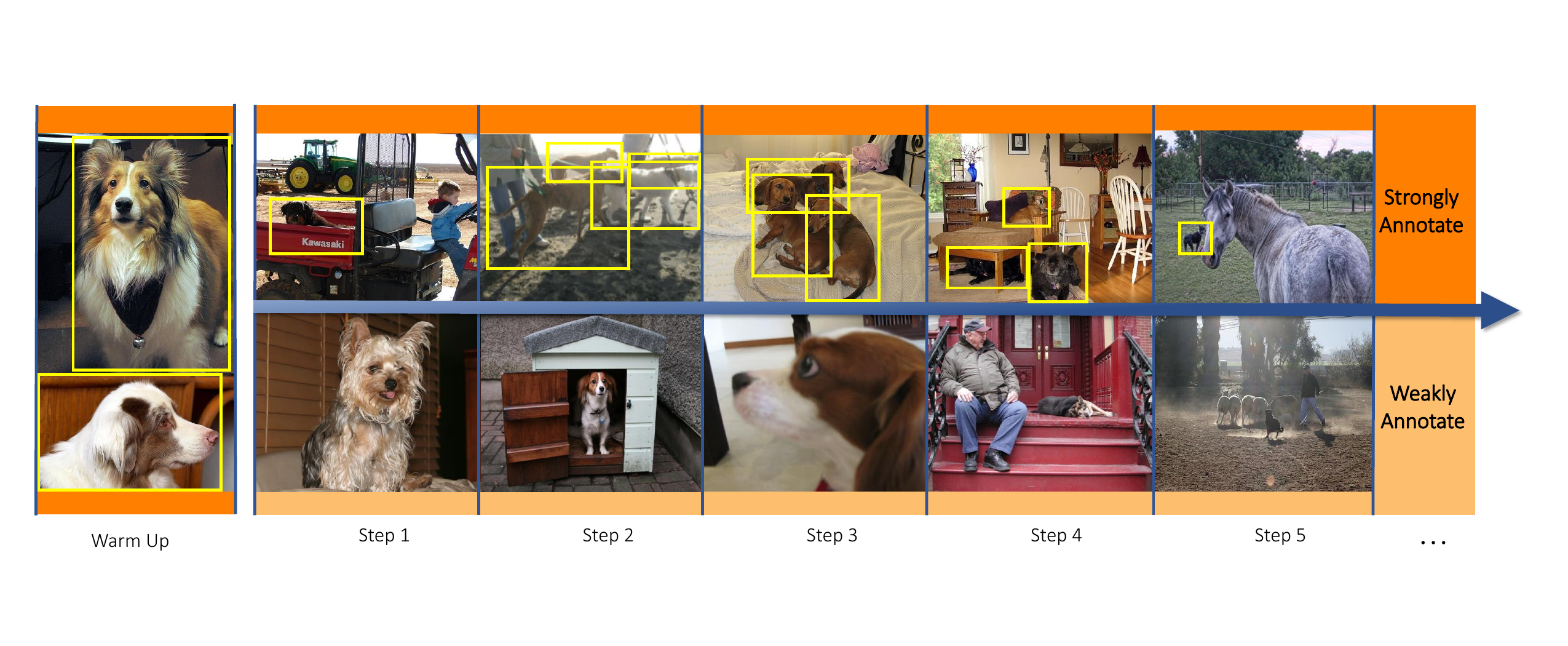}
\vspace{-38pt}
\caption{\small{\textbf{Visualization of the selected images in each step.} \textit{Left}: Two examples in the warm-up set which is fully annotated by $10\%$ budget. \textit{Up-Right}: Strongly annotated images per step. They are hard examples including occlusion, multiple instance or tiny scale. \textit{Bottom-Right}: Weakly annotated images per step. They are simple in the beginning but the difficulty increases when the detector is mature.
}} 
\vspace{-10pt}

\label{fig:vis}
\end{figure*}

This experiment simulates a real-world application of our method. We combine VOC07 and VOC12 data to simulate a larger pool of unlabeled images (called VOC0712) to choose from. As reported in Table \ref{tb:App}, when the budget is set at $87.2\%$ of the total budget (annotation budget for VOC07), we learn an object detector whose  performance is the same as an FSOD detector trained on the entire strongly labeled VOC07 training set. This is a budget saving of $12.8\%$. Now, if we choose to use  this total budget on VOC0712, our method  achieves a $73.0\%$ mAP, which is $2\%$ improvement over the  aforementioned FSOD detector on the same VOC07 test set. 
% \B{explain what it means to be 330\% budget} 
If all the 16551 images in the union of VOC07 and VOC12 set are strongly labeled, we use an extra $230\%$ of budget and only improve $3.4\%$ mAP.
% This result also elucidates that BAOD (active hybrid training) can reach approximately $95.5\%$ of the maximum performance possible when  VOC0712 is used, while only using a third of the budget needed to strongly annotate these two datasets. 

\begin{table}[h]
\caption{\small{\textbf{Simulation of a larger unlabeled image pool.} With $87.2\%$ budget, BAOD achieves the same performance as a detector trained on fully annotated VOC07. If the budget equals to the total budget of VOC07, BAOD achieves $2\%$ mAP improvement over FSOD with the same budget. Further annotating all the images can only improve $3.4\%$ mAP.}}
\vspace{-10pt}
\begin{center}
\label{tb:App}
\begin{tabular}{c||c|c|c|c}
\hline
\textbf{Train Set Pool} & \textbf{VOC07} & \multicolumn{3}{c}{\textbf{VOC0712} }\\  \hline \hline
\textbf{Max Budget}* & 100\% & \textbf{87.2\%} & 100\% & 330\% \\ \hline
\textbf{Final mAP}    & 0.710  & 0.710 & 0.730 & \textbf{0.764} \\ \hline
\end{tabular}
\end{center}
\begin{tablenotes} 
\small
\vspace{-10pt}
\item ~~~~ * The values are normalized by VOC07 trainval set cost.
\end{tablenotes} 
\vspace{-15pt}
\end{table}

\subsection{Easy Images and Weak Annotation First}
\label{S:Analyse}

We analyze the cost and number of images selected at every active selection step to investigate which type of training examples are more helpful in the sequential training process. Based on the final mAP of each category in VOC07  \cite{fast-rcnn}, we divide the twenty categories into three groups: Easy, Medium, and Hard.%describe how did we choose the categories} 
The left plot of Figure \ref{fig:analyse} illustrates that BAOD spends more budget to annotate \textbf{Easy} categories (shown in the green area) in the first several steps, while the cost of \textbf{Hard} categories (red area) increases when the detector becomes more accurate. These results align with concepts from curriculum learning, in which a larger number of \emph{easy} samples can be trained on first to bootstrap the model and then \emph{hard} samples are introduced progressively.

%indicates that introducing domain knowledge of curriculum learning can further increase the agent's behavior. 

The right plot of Figure \ref{fig:analyse} measures the percentage of selected images per active step that belong to various subsets of the data (combinations of annotation type chosen and difficulty). % the weakly and strongly annotated image numbers to see which kind of annotation is more dominant in certain active steps. 
Interestingly, the BAOD model tends to select more weak annotations at the beginning from all three groups of objects, since this kind of label is cheap and informative. Given a larger budget, the agent increases the proportion of strong annotations to further improve model performance. Note that there are no new weakly labeled images after active step 9 because all the images are already annotated. %However, we still need to spend extra budget to strongly annotate some images.
\vspace{5pt}

\subsection{Qualitative Results of the Active Selection}
\label{S:VIS}

Figure \ref{fig:vis} shows some selected strongly  and weakly labeled images based on  the BAOD experiment in section \ref{S:APP}. Each row of the images is from the \emph{dog} category in the VOC07 or VOC12 trainval set, and each column of images is selected at the same active step.

We show that in the first five active steps, strongly annotated images contain dog instances that are difficult to detect due to occlusion, multiple close instances or small-scale.
%\B{this contradicts the conclusion we made that Hard instances tend to be picked less in the beginning!! this whole paragraph needs to be rewritten or the strongly annotated images need to be changed} 
In contrast, the selected weakly annotated images during the same steps are relatively easier to locate. As the model gets mature the difficulty for both levels of annotations increases. For example, the image chosen in step 4 contains three small dogs, and the dogs appeared in step 5's images are small and black which makes them barely visible.

\section{Conclusion} \label{conclussions}
In summary, we introduce a novel budget-aware perspective to study the unexplored dimensions of the object detection problem. With a fixed budget, we compare both optimization and learn based sample methods to build diverse hybrid supervised object detection datasets which consist of both image level supervision and instance level supervision. The evaluation of  detectors that learned from these budget fixed datasets shows that the handcrafted optimization method on uncertainty score outperforms other general active learning methods including  random sampling, active learning, and reinforcement learning (shown in supplementary material).
With the optimal set-up, our proposed budget-aware approach can achieve the performance of a strongly supervised detector on PASCAL-VOC 2007 while saving $12.8\%$ of its original annotation budget. Furthermore, when  $100\%$ of the budget is used, our approach surpasses this performance by 2 percentage points of mAP.

\newpage
{\small
\bibliographystyle{ieee}
\bibliography{baod}
}

\onecolumn
\begin{appendices}
\section{Solutions to Optimization Functions}
\subsection{Image Sampling}
In section 3.1.1 and 3.1.2 of the submission, we propose two linear functions to approximate the mAP increment. Then we \textbf{maximize} it with several constraints. However, the proposed solution is an integer programming problem, and it cannot be solved in polynomial time. For example, the default solver uses Branch and Bound Algorithm (B\&B) and it takes more than 24 hours to find a global maximum. To solve this problem more efficiently, we take the relaxation of the original integer program shown as Eq.\ref{eq:relax}. Noted that the solution space of $\mathbf x_1, \mathbf x_2, \mathbf x_3$ becomes an interval $[0,1]$, but not a discrete set $\{0,1\}$, the relaxation Eq.\ref{eq:relax} can be solved in linear time. 
\begin{equation}\label{eq:relax}
\begin{array}{rrclcl}
\displaystyle \hat{\mathbf{x_1}},\hat{\mathbf{x_2}},\hat{\mathbf{x_3}} = \argmax_{\mathbf{x_1},\mathbf{x_2},\mathbf{x_3}\in [0,1]^{N}} & \multicolumn{3}{l}        
{\mathbf{s}^\top(\mathbf{x_1}+\mathbf{x_3}) 
+(\mathbf{\mu}- \mathbf{s})^\top\mathbf{x_2}} \\
\textrm{s.t.} 
&\displaystyle \mathbf{x_3} & \leq & \mathbf{\psi} \\
&\displaystyle \mathbf{x_1}+\mathbf{x_2} & \leq & \mathbf{1}-\mathbf{\psi} \\
&\displaystyle \mathbf{1}^\top (a  \mathbf{x_1}+b \mathbf{x_2}+c \mathbf{x_3}) &\leq& d \\ &\displaystyle \mathbf{1}^\top (a  \mathbf{x_1}+b \mathbf{x_2}+c \mathbf{x_3}) &\geq& d-a \\
% &\displaystyle a \mathbf{x_1}+b \mathbf{x_2}+c \mathbf{x_3} &\leq& d
\end{array}
\end{equation}
To project $\mathbf x_1, \mathbf x_2, \mathbf x_3$ back to the discrete set $\{0,1\}$, we take three float numbers $\epsilon_1,\epsilon_2,\epsilon_3$ as the thresholds of $\hat{\mathbf{x_1}},\hat{\mathbf{x_2}},\hat{\mathbf{x_3}}$, respectively. In another word,  every element in $\mathbf{x_k}$ larger than $\epsilon_k$ is set as $1$, otherwise it is $0$. 
\begin{equation}\label{eq:threshold}
\begin{array}{rcl}
\mathbf{x_1}^* &=& 1\{\hat{\mathbf{x_1}}>\epsilon_1\}\\
\mathbf{x_2}^* &=& 1\{\hat{\mathbf{x_2}}>\epsilon_2\}\\
\mathbf{x_3}^* &=& 1\{\hat{\mathbf{x_3}}>\epsilon_3\}
\end{array}
\end{equation}
%It is acceptable that we use an approximate global solution of the original problem, but t
The solution needs to be feasible after thresholding. More concretely, the first two constraints must be satisfied because we cannot give an invalid action (\eg It's impossible to annotate an image both weakly and strongly, but it might appear from the solution). 

We set $\epsilon_1=\epsilon,\epsilon_2=1-\epsilon,\epsilon_3=\epsilon$, where $\epsilon$ exhaustively goes from $0$ to $1$ to satisfy the last two budget constraints. In this case, it can be proven that the solution of Eq.\ref{eq:threshold} is also feasible from the original constraints.

\begin{proof}
(1) If $\psi(k)=0$, where $k$ can be the index of any element of vector $\hat{\mathbf{x_1}}$. 
Since $\hat{\mathbf{x_1}},\hat{\mathbf{x_2}},\hat{\mathbf{x_3}}$ is feasible, the first inequality in Eq.\ref{eq:relax} shows
\begin{equation}
    \hat{x_3}(k)\leq0\implies\hat{x_3}(k)=0.
\end{equation}
The second inequality in Eq.\ref{eq:relax} gives the inequality for $\hat{x_1}$ and $\hat{x_2}$.
\begin{equation}
    \hat{x_1}(k)+\hat{x_2}(k)\leq1\implies1-\hat{x_2}(k)\geq \hat{x_1}(k)\implies1\{1-\hat{x_2}(k)<\epsilon\}\leq 1\{\hat{x_1}(k)<\epsilon\}
\end{equation}
Here $1\{\}$ is the indicator function. When we apply Eq.\ref{eq:threshold} to check the two restrictions, the discrete solutions $\mathbf{x_1}^*,\mathbf{x_2}^*,\mathbf{x_3}^*$ are still feasible.
\begin{equation}
\begin{array}{rcccl}
     x_3^*(k)&=&1\{\hat{x_3}(k)>\epsilon\}=1\{0>\epsilon_3\}=0&=&\psi(k)  \\
     x_1^*(k)+x_2^*(k)&=&1\{\hat{x_1}(k)>\epsilon\}+1\{\hat{x_2}(k)>1-\epsilon\}\\
     &=&1\{\hat{x_1}(k)>\epsilon\}+1\{1-\hat{x_2}(k)<\hat{x_2}(k)\}\\
     &\leq&1\{\hat{x_1}(k)>\epsilon\}+1\{\hat{x_1}(k)<\hat{x_2}(k)\}\leq1&=& 1-\psi(k)
\end{array}
\end{equation}

(2) If $\psi(k)=1$, we have
\begin{equation}
    \begin{array}{rclcrcl}
         \hat{x_3}(k)&\leq&1    &\implies& \hat{x_3}(k)&=&1 \\
         \hat{x_1}(k)+\hat{x_2}(k)&\leq&0 &\implies& \hat{x_1}(k),\hat{x_2}(k)&=&0
    \end{array}
\end{equation}
Similar to previous discussion, we can have $x_3^*(k)\leq1$ and $x_1^*(k)=0,x_2^*(k)=0$ to fit the two constraints. 
\end{proof}

\subsection{Post Processing}
In the article, we give another optimization function to do post processing. This procedure checks the consistency between the image and instance-level annotations, and it removes abundant pseudo labels.

Given an image with the weakly annotation $\mathbf{\omega} \in \{0,1\}^{K}$, where $K$ is the number of categories. We assume the first detection model (\textit{teacher model}) gives $M$ positive predictions for the localization $P\in \mathbb{R}^{M \times 4}$, classification $A\in \{0,1\}^{M \times K}$,
and the positive class confidence vector $\mathbf{q}\in [0,1]^M$. Eventually, the post processing returns a sparse $M$-dimensional binary vector $\mathbf{y}$ as Eq.\ref{eq:PLM}, where $\alpha=0.3,\beta=3$.
\begin{equation} \label{eq:PLM}
\begin{array}{rrclcl}
\displaystyle \min_{\mathbf{y}\in \{0,1\}^{M}} & \multicolumn{3}{c}{-\mathbf{y}^\top \mathbf{q}} \\
\textrm{s.t.} & \mathbf{y}^\top A (\mathbf{1}-\mathbf{w})& = & 0 \\
& \forall_{{y}_i,{y}_j=1} IoU(P_i, P_j) &\leq&\alpha\\
&\displaystyle |\mathbf{y}|_0 & \in & [1,\beta]  
% &\displaystyle |\mathit{x}|_0 & \leq & K 
\end{array}
\end{equation}
In this  problem, our objective function takes high confidence predictions by choosing $\mathbf{y}\in \{0,1\}^{M}$. In the first constraint, $\mathbf{y}^\top A$ accumulates the selected predictions by their categories. The following inner product with $(\mathbf{1}-\mathbf{w})$ returns zero only when all the predicted classes are in the weak annotation. On the other hand, the second constraint removes the heavily overlapped predictions while the third one makes the binary vector $\mathbf{y}$ sparse. We develop Alg.\ref{alg:clean} to solve the above optimization.
\begin{center}

\begin{minipage}{.7\linewidth}
\begin{algorithm}[H]
    \SetKwInOut{Input}{Input}
    \SetKwInOut{Output}{Output}

    \Input{Weak annotation $\mathbf{\omega} \in \{0,1\}^{K}$, localization $P\in \mathbb{R}^{M \times 4}$, classification $A\in \{0,1\}^{M \times K}$, the positive class confidence for $M$ predictions $\mathbf{q}\in [0,1]^M$, $\alpha$, $\beta$;}
    \Output{a binary vector $\mathbf{y}\in \{0,1\}^{M}$;}
    $\mathbf{y}=\mathbf{1} $;
    
    \For{i=1:M}
      {
        if not $A(i,:) < \mathbf{\omega}$, $y(i)=0$;
      }
    $\mathbf{y'}$ = NMS index for [P,A,$\mathbf{q}$], do it by class with threshold $\alpha$
    
    $\mathbf{y} = \mathbf{y} .* \mathbf{y'}$
    
    \If{$sum(\mathbf{y})>\beta$}{assign the top $\beta$ remained predictions to $y$}
    \caption{Noise Cleaning}\label{alg:clean}
\end{algorithm}
\end{minipage}

\end{center}

\section{More Budget-Aware Methods and Experiments}
\subsection{Sample from Small Uncertainty Images}
In the Section 3.1 of the submission, to apply Uncertainty Sampling (US) method, we sort images in \textbf{descending} order of uncertainty and choose high uncertainty images. Moreover,
we also conducted an experiment in which images were sorted by the uncertainty score in an \textbf{ascending} order, and we pick low uncertainty images in every step. The other settings are the same to US's. Table \ref{tb:EXP1} compares these two methods.
\begin{table}[h]
\caption{\small{\textbf{Budget-Average mAP using Largest and Smallest Uncertainty Scores.} When the budget is low (10\%-30\%), collecting images with small uncertainty score (shown as SUS) is more effective. However, when the dataset size is large enough, choosing large uncertainty scores (shown as LUS) is preferable. }}
\begin{center}
\small
\label{tb:EXP1}
\begin{tabular}{c||c|c|c|c}

\hline
\textbf{Selection Method} & \multicolumn{2}{c|}{\textbf{FSOD}} & \multirow {2}{*}{\textbf{RL}} & \textbf{Opt.}\\  \cline{1-3} \cline{5-5}
\textbf{Sampling Function} &  \textbf{LUS} & \textbf{SUS} &  & \textbf{US} (BAOD) \\ \hline \hline
\textbf{Low Budget Range}    & 52.3  & {53.1} & 56.6 & 57.1\\ \hline
\textbf{Mid Budget Range}    & {63.1}  & 62.8 & 64.1 & 66.0\\ \hline
\textbf{High Budget Range}   & {68.6} & 67.7 & 68.5 & 69.5\\ \hline
\end{tabular}
\end{center}
\end{table}

When the budget is in low range ($10\%-30\%$), collecting images with small uncertainty score is more effective. However, when the dataset size is larger, choosing large uncertainty scores is preferable. 
This behaviour follows the conclusions of Bengio \etal \cite{bengio2009curriculum}, who suggested that it is better to train models using curriculum learning, \textit{i.e.} giving \emph{easy} images to the model in  early training and then giving  \emph{hard} images once the model has learned better representations. 
This experiment also shows that there are better solutions to the BAOD problem. For example, we can find a better sampling function which also considers the current state of the detection model. Assembling the sampling function into the optimization selection method will produce a better budget-mAP performance.

\subsection{Reinforcement Learning}
\subsubsection{Method}

The problem of finding the next active batch of training examples can be naturally formulated as a reinforcement learning problem. This approach allows us to learn a strategy directly from the trial-and-error experience and to avoid the specific mathematical model of the previous sections. The reinforcement learning methods do not need any prior knowledge about the environment, but they can still construct an optimal strategy. Thus, it is easily extensible to stochastic environments with variable set-ups such as strongly-weakly annotation costs ratio and a total amount of budget. 

Suppose that we already have a training set, and we regress a detector with performance $m_0$. 
In the next active step, we evaluate the uncertainties of a subset of unlabeled images. The active learning agent interacts with the environment in the form of the
\textbf{difficulty preference} and \textbf{annotation type}. We use $O_t$ as defined in section 3.1.2 of the submission as the observed state, which considers both the classification and localization performance. The agent returns two actions from current state: (1) asking for low/medium/high uncertainty images and (2) giving a strongly/weakly annotation. In conclusion, the RL agent gives a choice $\mathit{a}$ from \textit{three by two} possible actions matrix. 

In the RL training process, we take the detection model increment performance $\Delta_t$ as a reward. If there is still enough budget, the agent chooses the next active batch from the new observed state $O_{t+1}$. If we run out of budget after $T$ active steps, the total reward is $R = \sum_{t=1}^{T} \Delta_t=m_T-m_0 $. Here $m_0$ is the performance of a randomized warm-up model with a fixed seed. By maximizing the return $R$, the agent learns a policy $\pi$ that maximizes the final model performance $m_T$. 

To train the agent, we learn a policy with $Q$-learning\cite{watkins1992q} which learns to
approximate $Q$-function $Q^\pi (\mathit{a}, O_t)$ with a three-layer neural network. Given an action $\mathit{a}$ and the current state $O_t$, it  returns the expected value from policy $\pi$.

\subsubsection{Experiment}

At every active step, the $Q$ agent gives insights on which images to label and which type of annotations is required. Different from the previous pipeline, the RL method does not evaluate all the unlabeled images, but renders a preference for easy, middle or hard images. This significantly expedites the training process. We firstly compute the number of required images from the budget and annotation types, then we stop evaluating unlabeled images' uncertainties, once we get enough images. Table \ref{tb:EXP1} shows that RL performs better than FSOD, since it learns a policy to weakly or strongly annotate images and takes advantage of hybrid training. Nevertheless, it performs worse than US and LAL optimization. The reinforcement learning environment is built by VOC12 training set and validation set. We simulate the active learning process on four Titan Xp GPUs for five days to train the $Q$-learning agent, and the agent is used to solve the budget aware object detection problem on VOC07 trainval set and test set.

\section{Definition of Easy, Medium and Hard Classes}
We analyze the cost and number of images selected at every active selection step to investigate which type of training examples is more helpful in the sequential training process. Based on the final mAP of each category in VOC07, we divide the twenty categories into three groups: Easy, Medium and Hard. The mapping is motivated and shown in Table 2.
\begin{table}[h]
\caption{\small{\textbf{The mapping from twenty categories in PASCAL VOC to Easy, Medium and Hard classes}. This mapping is based on the final mAP of each category. }}
\vspace{-5pt}
\begin{center}
\setlength{\tabcolsep}{2pt}
\small
\label{tb:mapping}
\begin{tabular}{|r|c|c|c|c|c|c|c|c|c|c|c|c|c|c|c|c|c|c|c|c|}

\hline \textbf{Class} &
\multicolumn{8}{c}{\textbf{Easy}} &
\multicolumn{6}{|c|}{\textbf{Medium}}&
\multicolumn{6}{c|}{\textbf{Hard}} \\ \hline
\textbf{mAP$^*$} & 88.9&88.4& 88.1& 87.6& 86.6&86.3&85.9&84.3&
82.3& 82.0&80.4&80.1&78.9&77.7&
75.8 &65.7& 68.9&70.8 & 63.6& 53.6\\ \hline
\textbf{Category} & cat & car & bus & horse&train&  cow& dog& areo& 
person& bike& sheep& mbike&  tv& bird&
sofa & table &  boat& bottle & chair & plant \\ \hline
\multicolumn{13}{l}{*  data is from the last setup of Table 6 in the report \cite{jjfaster2rcnn}.}
\end{tabular}
\end{center}
\vspace{-5pt}

\end{table}

\section{Discussion on Annotating Time}
In the submission, we assume $a= 34.5$ or $a= 7$, and $b= 1.6$. We explain these values in this section.

First of all, annotating an image does not take a fixed time. From the technique used in ImageNet\cite{imagenet}, the \textbf{Median} of annotating a bounding box is $25.5s$, but an extra of $9.0s$ and $7.8s$ is required to control its quality. However, using the state-of-the-art method \emph{extreme clicking for efficient object annotation} \cite{papadopoulos2017extreme} can reduce the strongly annotating time to $7s$. To consider most bounding box annotation methods, we assume this value to be $7s$, $21s$ or $34.5s$.
On the other hand, answering a binary question takes around $1.6$ seconds\cite{papadopoulos2016we}, but answering twenty binary question does not take $1.6\times20=32s$.  \cite{papadopoulos2016we} finds that clicking on one of five on-screen buttons takes around $2.4$ second. In the original ImageNet paper, people select a sequence of queries for humans such that we achieve the same labeling results with only a fraction of the cost of the naive approach.
It is more efficient to organize the semantic concepts into hierarchies \cite{thorpe1996speed}. More importantly images are collected from the internet through keyword search, which implies that many image level labels are free. % For those reasons, the weakly labelled annotation cost is assumed to be $1.6s$.

In our Budget-Aware Object Detection problem, we only care about \textbf{budget ratios} between strong annotations and weak annotations, although there are multiple instances in one image. For this reason, we have experiments with different ratios to compensate the cost assumptions and to show that at different ratios our method still works. In the last chapter of the supplementary material, we have a table to show the experiments with different cost ratios (CR) in $\{34.5:1.6, 21:1.6, 7:1.6\}$ of strong and weak annotations.

\section{Budget-mAP Curves}
We present the original Budget vs mAP figures for Table 1,2 and 3 in the submission, respectively. The curves shown here are mAP scores before integration, which are more detailed but noisy. Please check section \ref{S:raw} for the table representation.

\begin{figure}[h]
\begin{minipage}[t]{.33\textwidth}
\centering
\includegraphics[width=\linewidth]{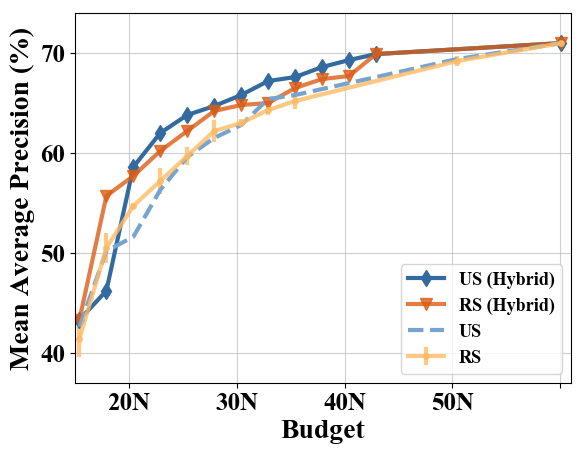}
\end{minipage}
\begin{minipage}[t]{.33\textwidth}
\centering
\includegraphics[width=\linewidth]{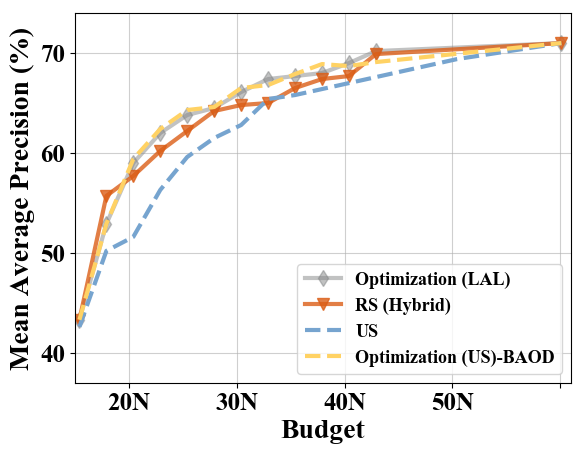}

\end{minipage}
\begin{minipage}[t]{.33\textwidth}
\centering
\includegraphics[width=\linewidth]{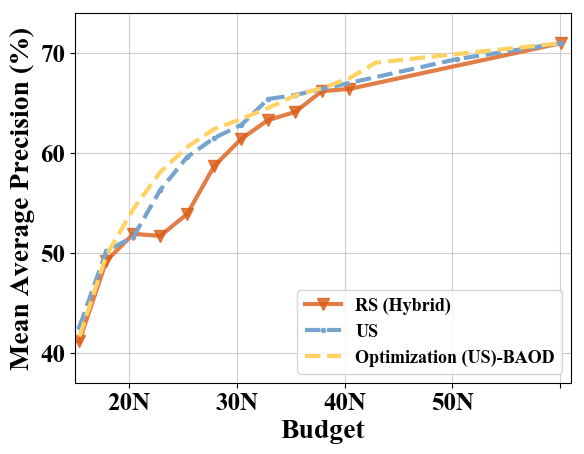}
\end{minipage}
\caption{\small{
\textit{left}: \textbf{Budget-mAP curves using fully and hybrid
training pipelines.} Orange bars compare the two training
pipelines using random sample while Blue bars show both
pipelines using US sampling.
\textit{Middle}: \textbf{Budget-mAP curves using FSOD, hybrid
training, and optimization methods.} Blue bars show US
using FSOD. Orange bars show RS hybrid baseline. Gold
bars show US Optimization (BAOD). Dark Grey Bars
show LAL Optimization.
\textit{right}: \textbf{Budget-mAP curves using a lower cost for
strong annotations}. Blue Bars represent FSOD with US
sampling. Orange Bars represent Hybrid training with RS
sampling. Gold Bars represent US Optimization. All the
methods use a smaller gap between the weak and strong
annotation costs.} }
\end{figure}

\newpage
\section{Raw Experiment Data}\label{S:raw}
\begin{table}[!h]
\setlength{\tabcolsep}{3.5pt}
\caption{\small{\textbf{Table Representation for Raw Experiment Data}}}
\vspace{-5pt}
\centering
\begin{tabular}{r|cc|ccccccccc|cccc|c}
\hline
& & &\multicolumn{13}{c}{Budget Percentages} & Full\\
\hline
&H$^*$&CR$^{**}$& 10.80 & 15.79 & 20.77 & 25.76 & 30.75 & 35.74 & 40.73 & 45.72 & 50.71 & 55.70& 60.69 & 65.68 & 80.70 & 100 \\\hline
RS&\xmark&${H}$ & 0.414&0.505&0.547&0.572&0.597&0.622&0.630&0.643&0.652&-&-&-&0.692&0.71  \\
US&\xmark&${H}$  &      0.430&0.506&0.519&0.552&0.599&0.621&0.623&0.651&0.657&-&-&-&0.695&0.71   \\
SUS&\xmark&${H}$  & 0.397&0.502&0.546&0.575&0.602&0.62 &0.633&0.636&0.648&-&-&-&0.682&0.71  \\ 
LAL&\xmark&${H}$  & 0.429&0.495&0.545&0.569&0.58 &0.606&0.634&0.64 &0.644&-&-&-&0.692&0.71 \\\hline
RS&\cmark&${H}$ & 0.408&0.562&0.576&0.608&0.612&0.637&0.651&0.655&0.663&-&-&-&-&0.71  \\
US&\cmark&${H}$ & 0.433&0.462&0.586&0.62 &0.638&0.647&0.658&0.672&0.676&-&-&-&-&0.71  \\
\hline
OPT-US&\cmark&${H}$& 0.433&0.529&0.594&0.624&0.643&0.646&0.665&0.668&0.679&0.689&0.687&0.691&-&0.71 \\
OPT-LAL&\cmark&${H}$ &0.389&0.529&0.59 &0.62 &0.638&0.645&0.661&0.674&0.677&-&-&-&-&0.71  \\
\hline
US&\cmark&${L}$ & 0.412 & 0.498 & 0.509 & 0.527 & 0.573 & 0.613 & 0.626 & 0.653 & 0.664 & 0.675 & 0.673 & - & - & 0.71  \\
RS&\cmark&${L}$ & 0.412 & 0.492 & 0.519 & 0.517 & 0.539 & 0.587 & 0.614 & 0.633 & 0.641 & 0.662 & 0.664 & - & - & 0.71  \\

OPT&\cmark&${L}$  & 0.42 &0.518&0.548&0.586&0.609&0.62 &0.64 &0.648&0.665&0.67 &0.683&0.689&-&0.71 \\
\hline

RS&\cmark&${M}$  & 0.443&0.554&0.576&0.6  &0.62 &0.62 &0.654&0.652&0.655&0.672&0.675&0.689&-&0.71   \\

OPT&\cmark&${M}$  & 0.433&0.544&0.582&0.608&0.632&0.643&0.652&0.664&0.677&0.684&0.687&0.693&-&0.71  \\
\hline
\hline
Budget&-&-&10.8&20.8&30.8&40.8&50.8&60.8&70.8&80.8&90.8&92.8&96.8&-&-&-\\ \hline
VOC0712 &\cmark&${M}$ 
& 0.356&0.566&0.611&0.641&0.674&0.672&0.696&0.701&0.715&0.717 &0.72  &-&-&0.73\\
\hline
Budget &-&-&11.09&13.62&18.1 &22.58&27.06&31.53&36.15&40.49&44.97&49.45&53.93 &-&-&-\\ \hline
RL &\cmark&${H}$&0.433&0.525&0.561&0.579&0.603&0.615&0.634&0.645&0.648&0.657&0.664 &-&-&0.71\\ \hline
\multicolumn{13}{l}{* H for Hybrid Learning}\\
\multicolumn{13}{l}{** CR for cost ratio of strong and weak annotations}
\end{tabular}
\end{table}

\newpage
\section{More Visualizations}

Below we show some visualization figures of strongly labelled and weakly images for each of the classes chosen by US based Optimization. Each column indicates one step of the active learning pipeline and each of the rows represents a class. The warm up set is represented on the first column showing only strongly labeled images chosen randomly. For the next 5 steps we can observe on the top sub-row strongly labeled images and on the bottom sub-row weakly labeled images. The bounding boxes shown in this figure are not the original annotations from PASCAL, they were drawn manually only for this visualization purposes.

\newcolumntype{M}[1]{>{\centering\arraybackslash}m{#1}}

    \centering
    \begin{tabularx}{\linewidth}{@{}lX@{}}
    Class 1 & \includegraphics[width=\linewidth]{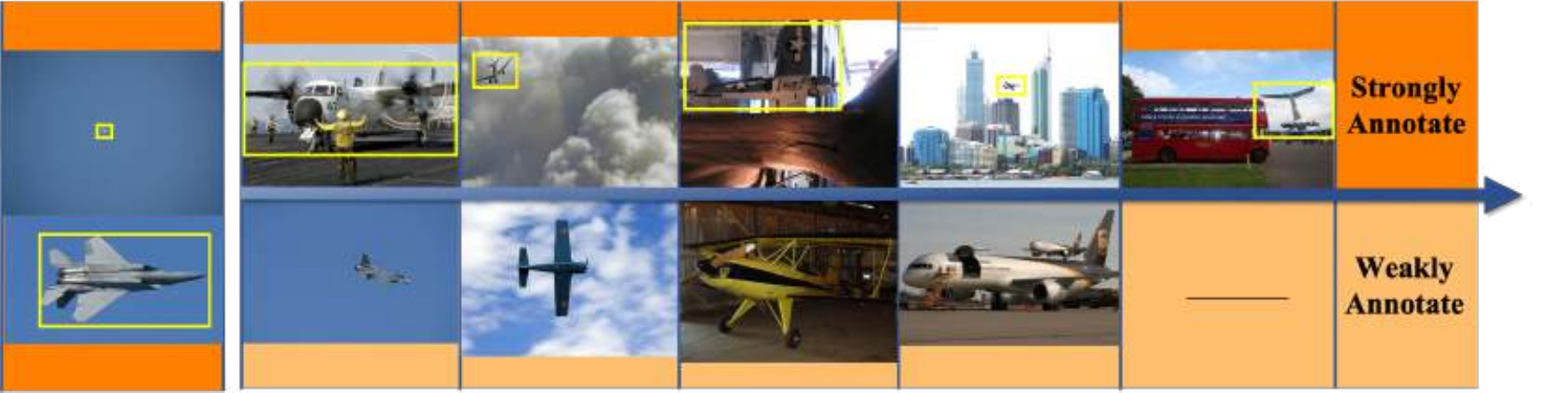}  \\
    Class 2 & \includegraphics[width=\linewidth]{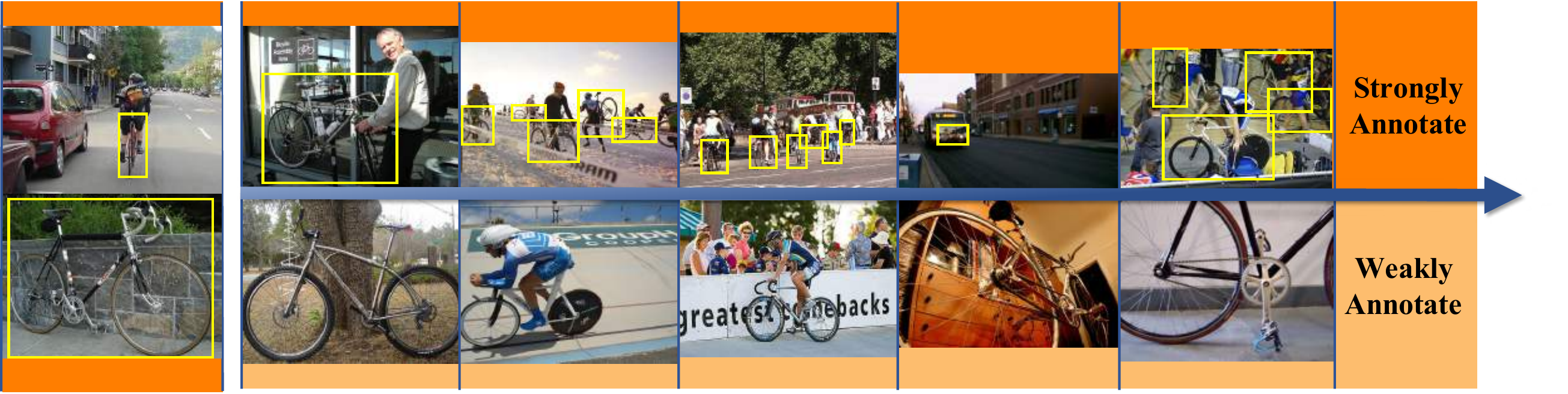}  \\
    Class 3 & \includegraphics[width=\linewidth]{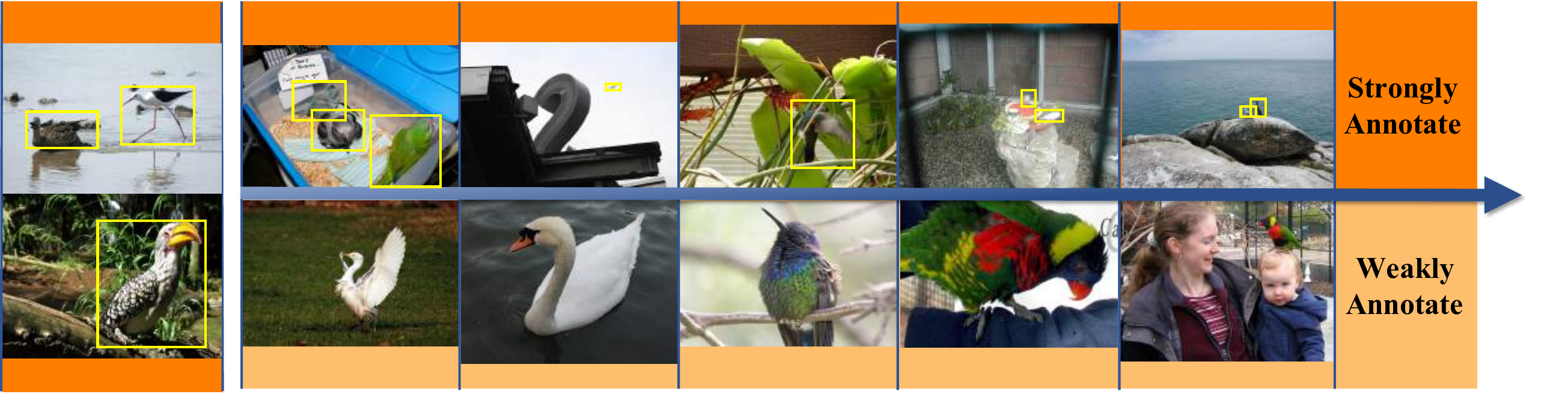}  \\
    Class 4 & \includegraphics[width=\linewidth]{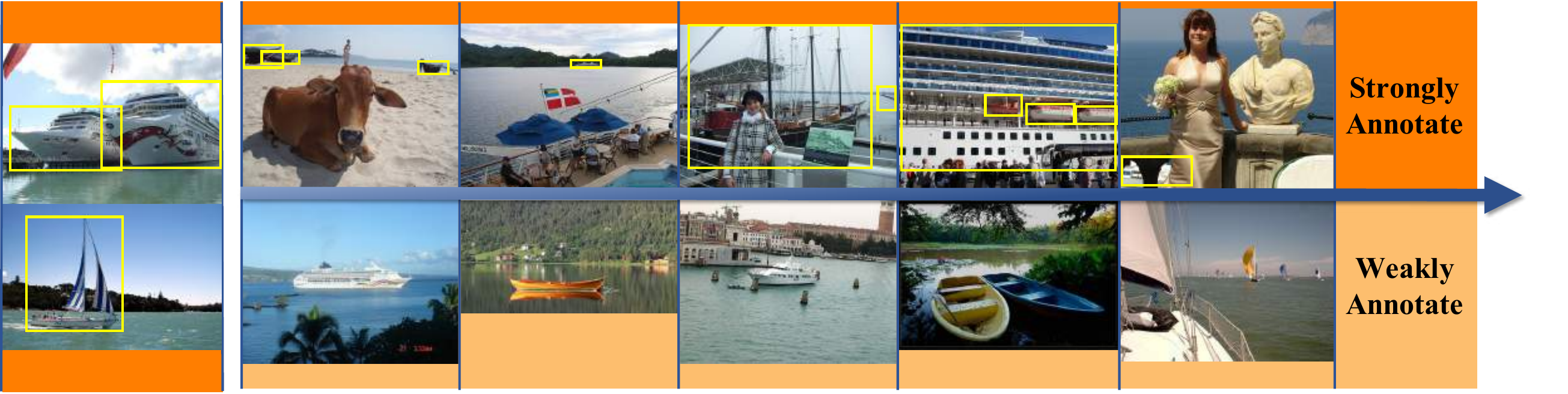}  \\
    Class 5 & \includegraphics[width=\linewidth]{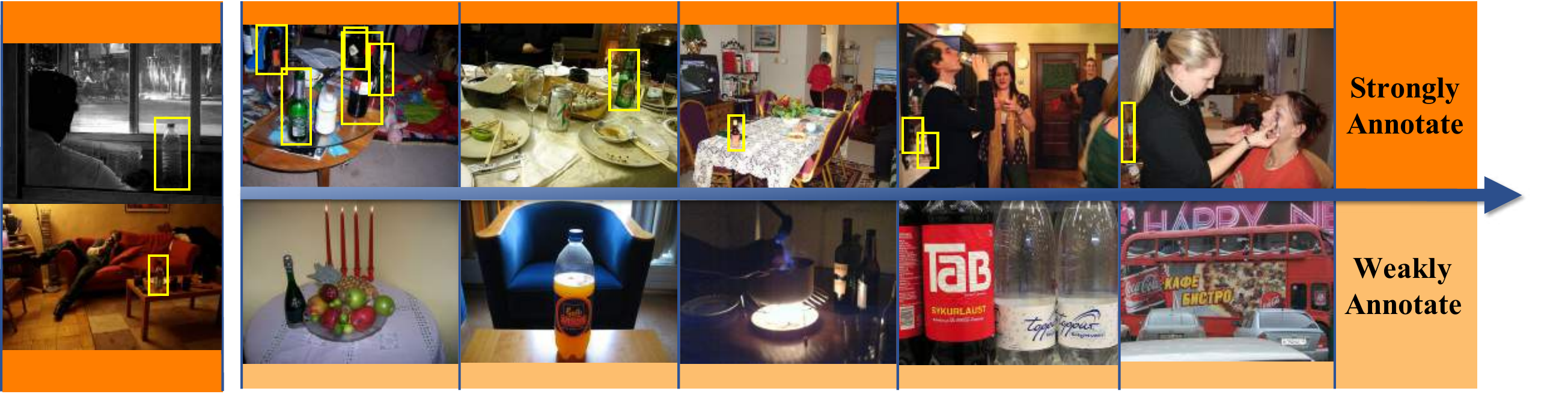}  \\
    Class 6 & \includegraphics[width=\linewidth]{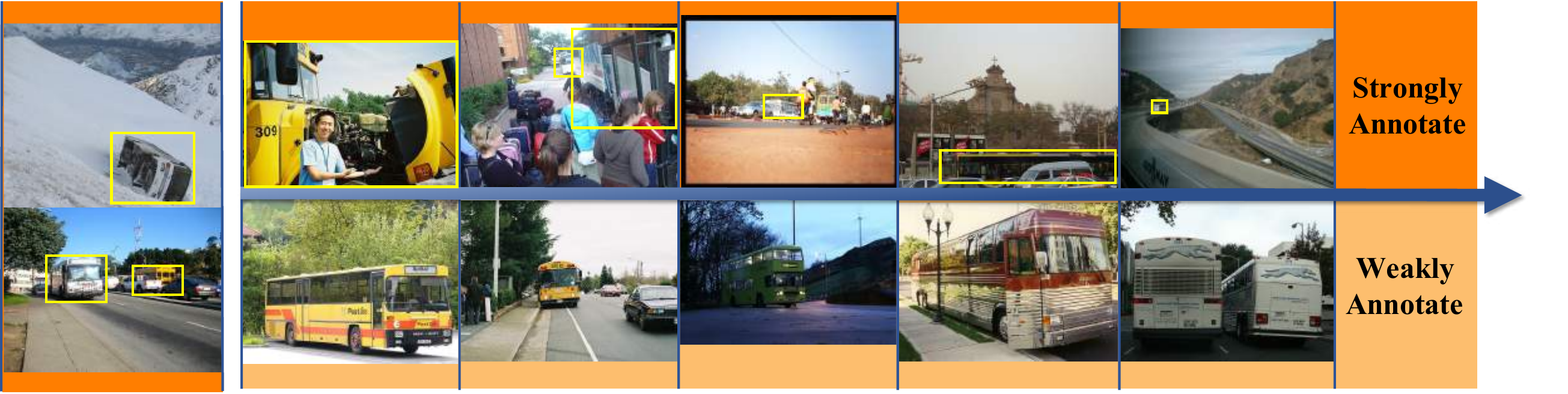}  \\
    Class 7 & \includegraphics[width=\linewidth]{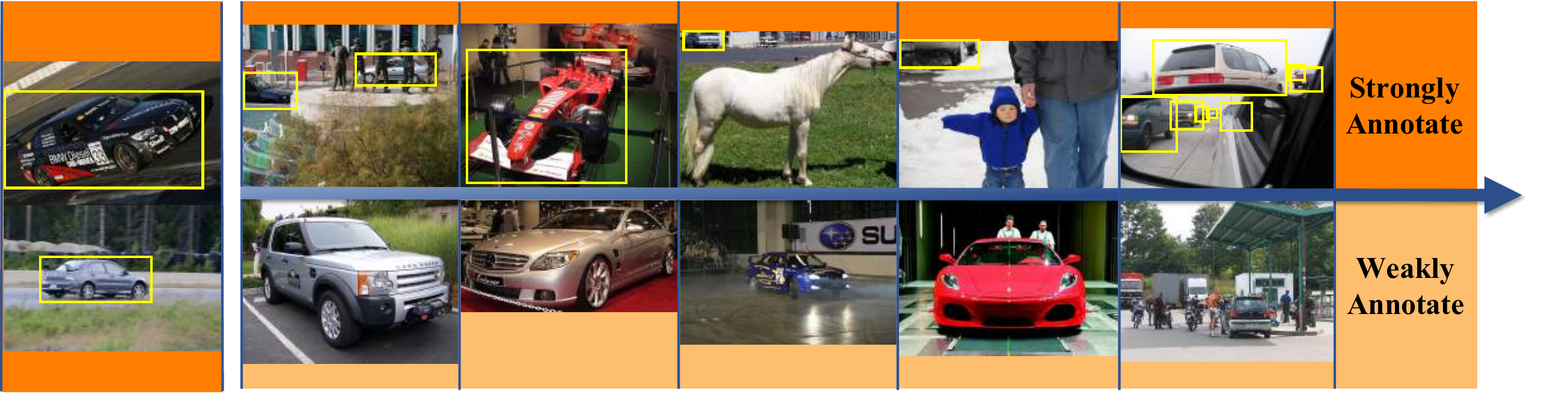}  \\
    Class 8 & \includegraphics[width=\linewidth]{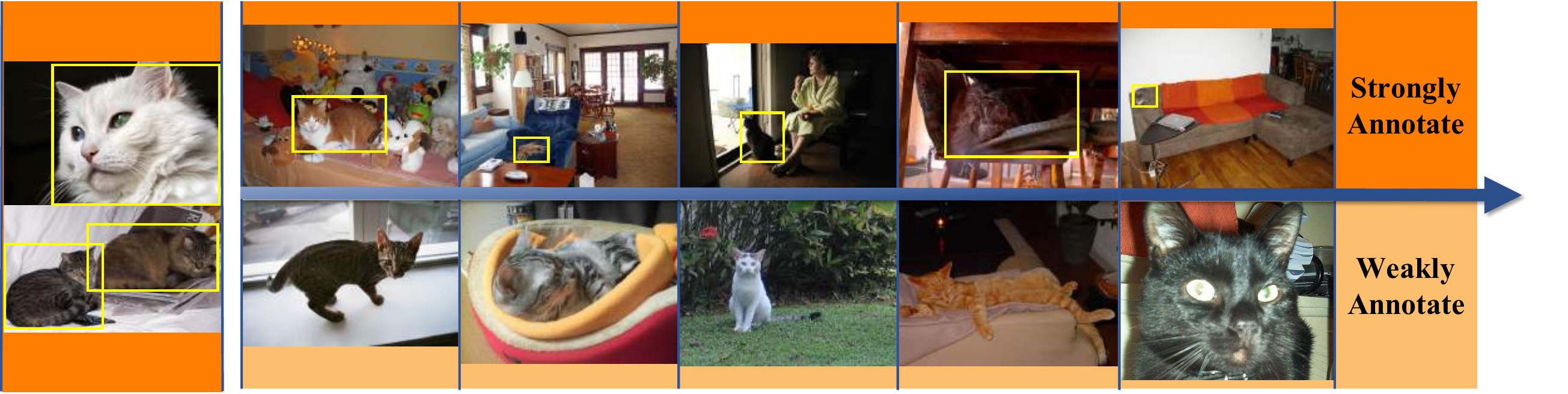}  \\
    Class 9 & \includegraphics[width=\linewidth]{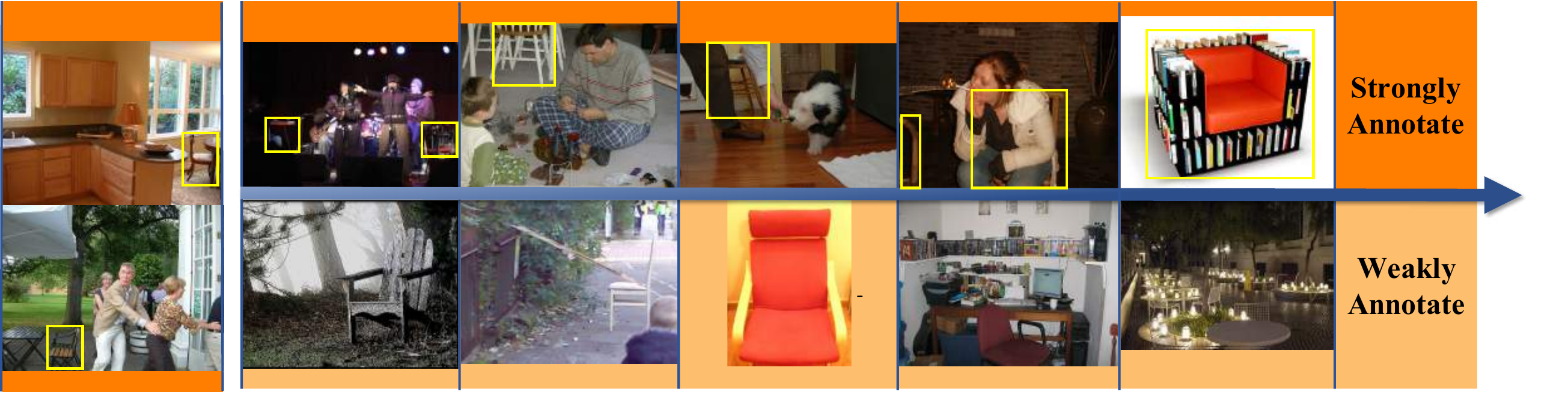}  \\
    Class 10 & \includegraphics[width=\linewidth]{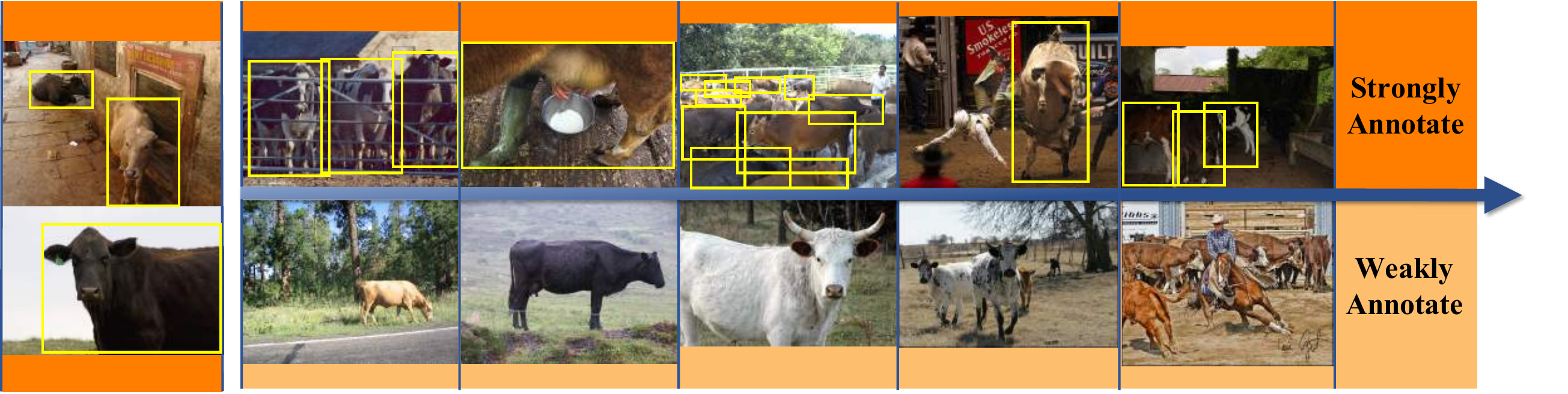}  \\
    Class 11 & \includegraphics[width=\linewidth]{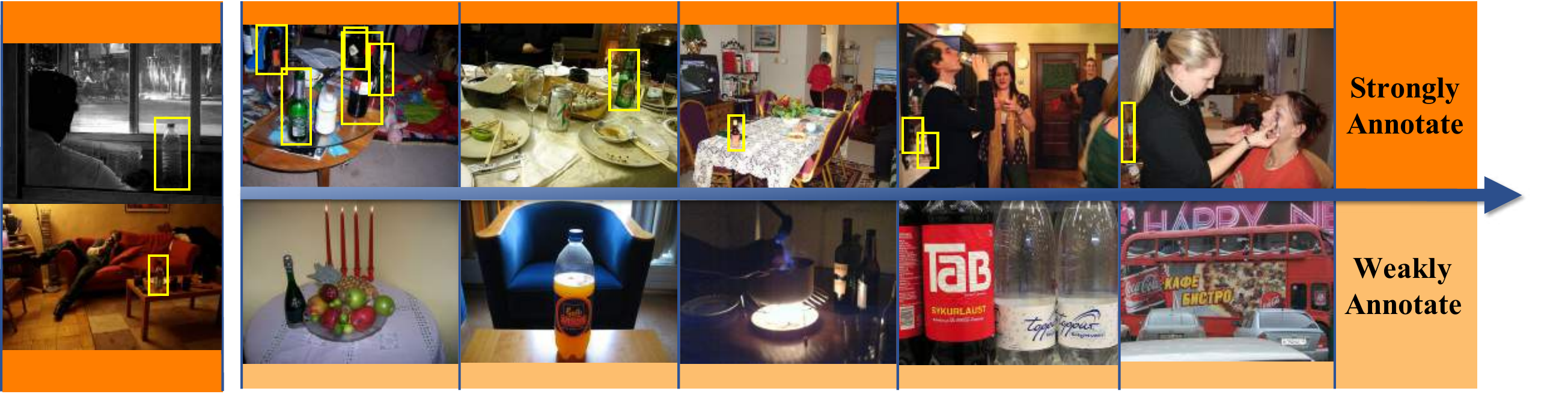}  \\
    Class 12 & \includegraphics[width=\linewidth]{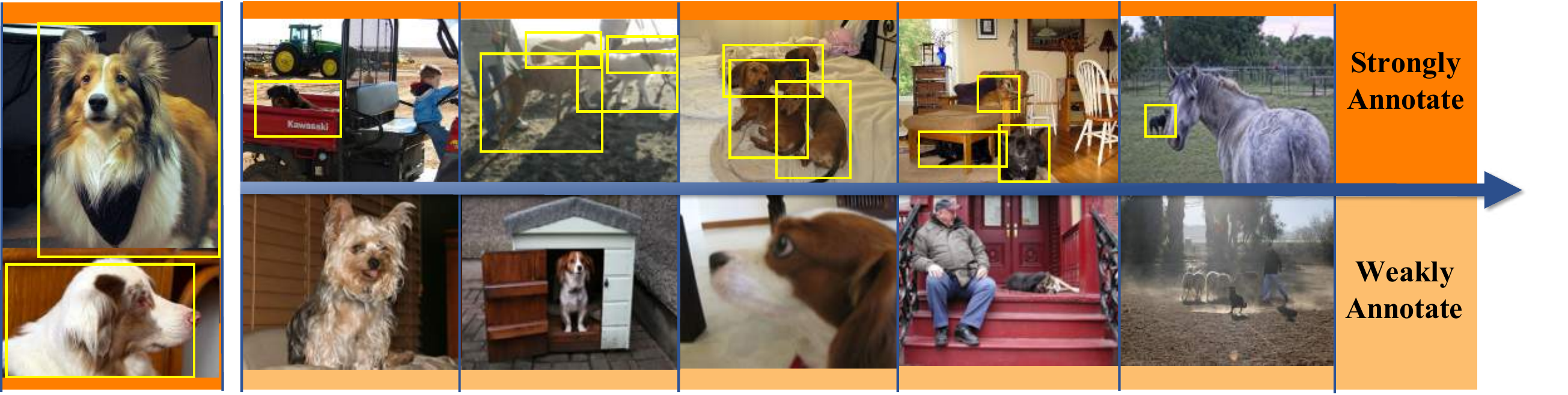}  \\
    Class 13 & \includegraphics[width=\linewidth]{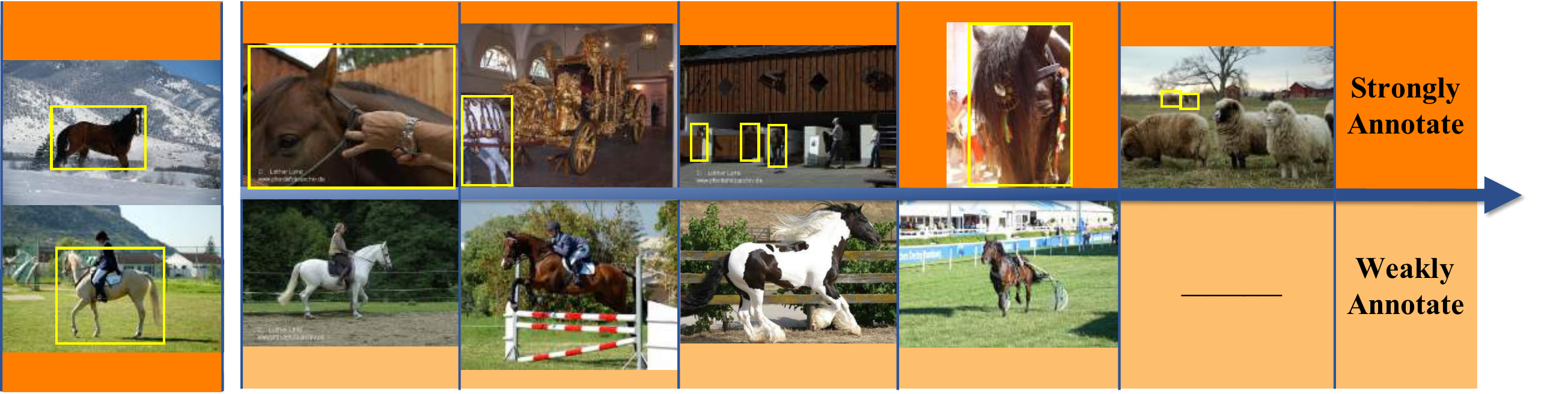}  \\
    Class 14 & \includegraphics[width=\linewidth]{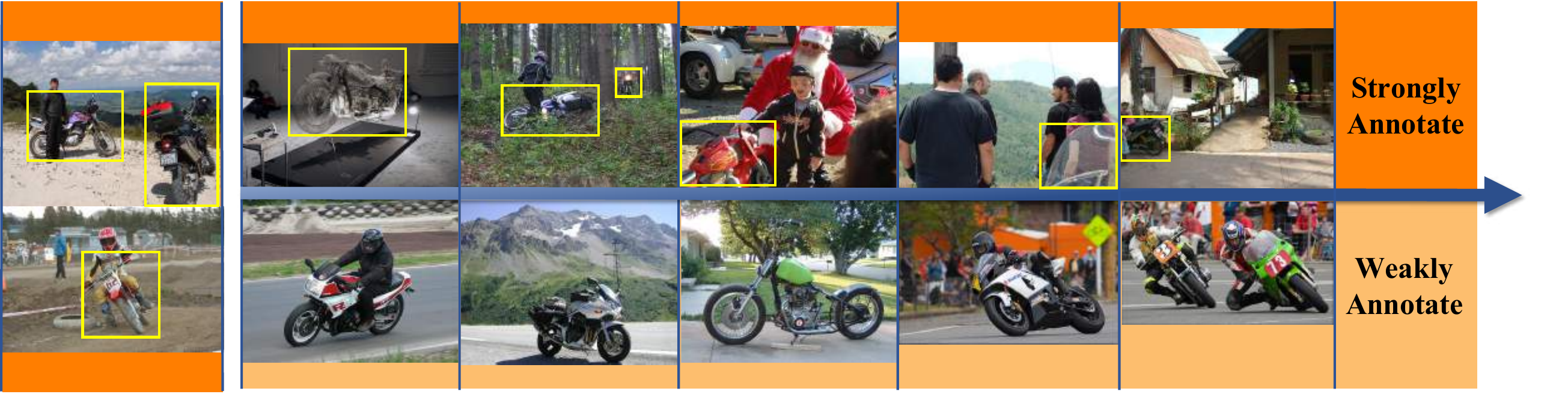}  \\
    Class 15 & \includegraphics[width=\linewidth]{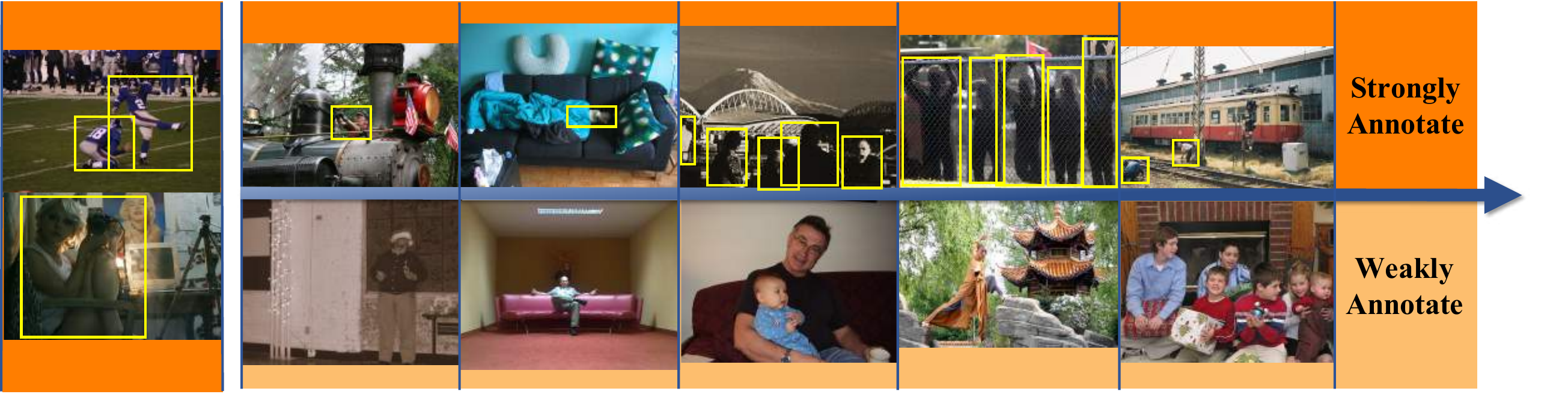}  \\
    Class 16 & \includegraphics[width=\linewidth]{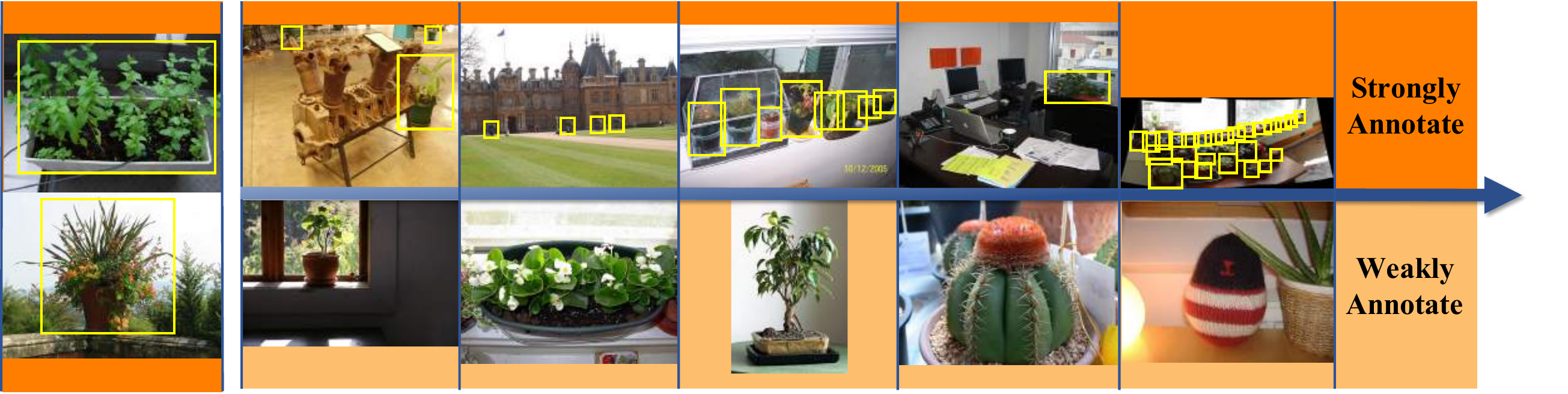}  \\
    Class 17 & \includegraphics[width=\linewidth]{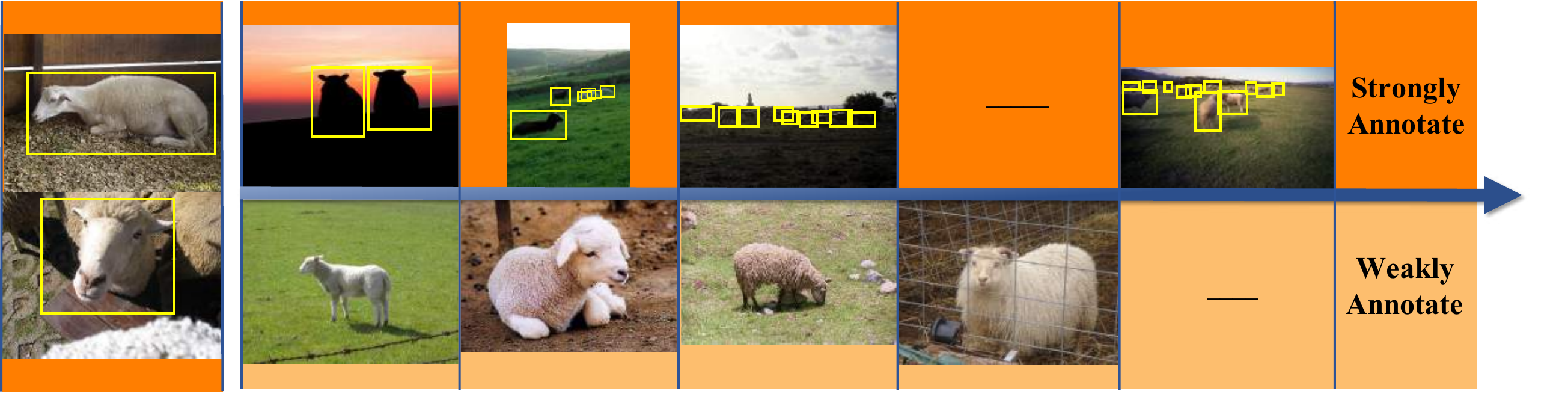}  \\
    Class 18 & \includegraphics[width=\linewidth]{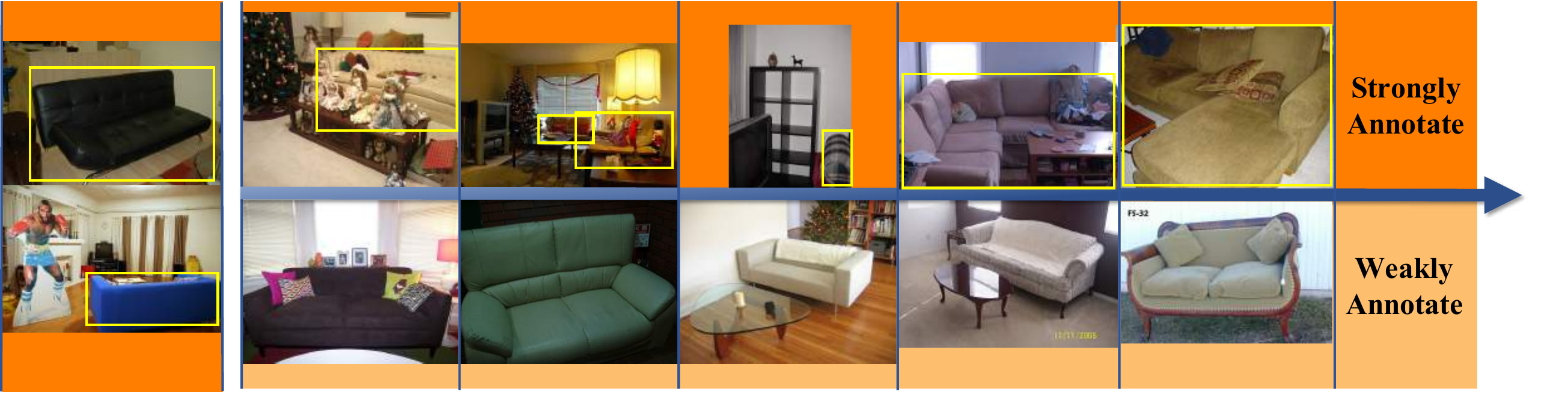}  \\
    Class 19 & \includegraphics[width=\linewidth]{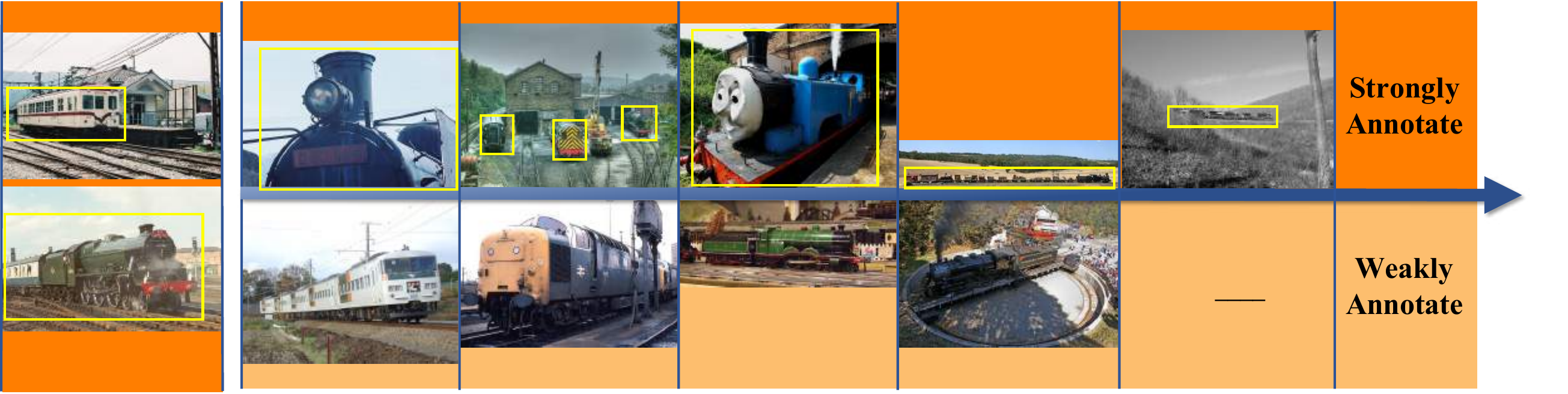}  \\
    Class 20 & \includegraphics[width=\linewidth]{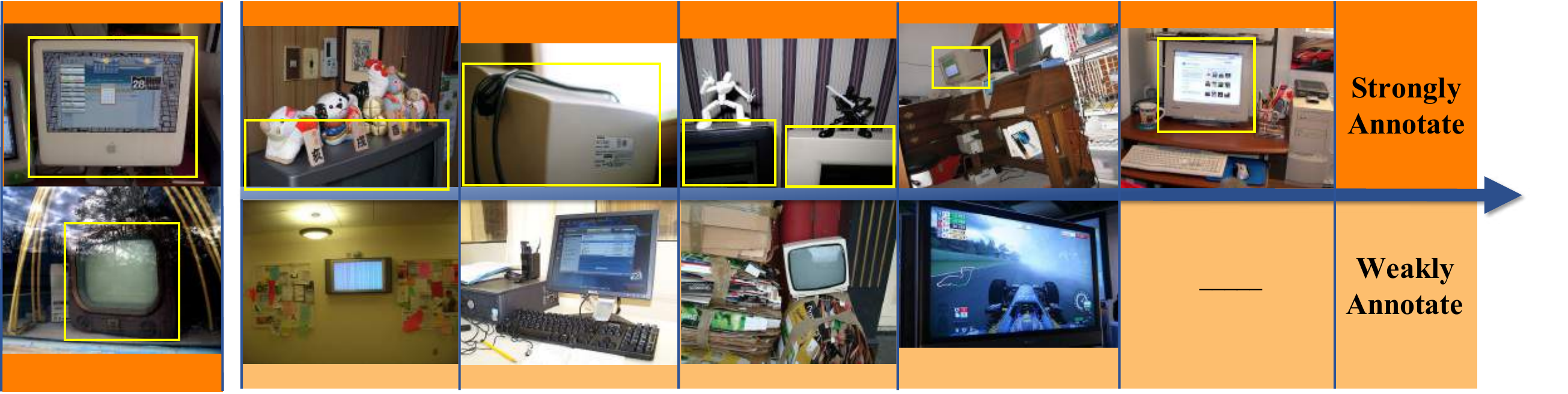}  \\
\caption{\small{\textbf{Visualization of the results for each of the 20 classes of PASCAL VOC}. Each column indicates one step of the active learning pipeline and each of the rows represents a class. Step 0 is represented on the first column showing only strongly labeled images chosen randomly. For the next 5 steps we can observe on the top sub-row strongly labeled images and on the bottom sub-row weakly labeled images. The bounding boxes shown in this figure are not the original annotations from PASCAL, they were drawn manually only for this visualization purposes.}}
\label{tbl:table_of_figures}
\end{tabularx}

\newpage
{\small
\bibliographystyle{ieee}
\bibliography{baod}

\begin{thebibliography}{10}\itemsep=-1pt

\bibitem{bai2018finding}
Y.~Bai, Y.~Zhang, M.~Ding, and B.~Ghanem.
\newblock Finding tiny faces in the wild with generative adversarial network.
\newblock {\em CVPR}, 2018.

\bibitem{beluch2018power}
W.~H. Beluch, T.~Genewein, A.~N{\"u}rnberger, and J.~M. K{\"o}hler.
\newblock The power of ensembles for active learning in image classification.
\newblock In {\em Proceedings of the IEEE Conference on Computer Vision and
  Pattern Recognition}, pages 9368--9377, 2018.

\bibitem{bengio2009curriculum}
Y.~Bengio, J.~Louradour, R.~Collobert, and J.~Weston.
\newblock Curriculum learning.
\newblock In {\em Proceedings of the 26th annual international conference on
  machine learning}, pages 41--48. ACM, 2009.

\bibitem{9}
H.~Bilen, M.~Pedersoli, and T.~Tuytelaars.
\newblock Weakly supervised object detection with convex clustering.
\newblock In {\em CVPR}, pages 1081--1089, 2015.

\bibitem{12}
H.~Bilen and A.~Vedaldi.
\newblock Weakly supervised deep detection networks.
\newblock In {\em CVPR}, pages 2846--2854, 2016.

\bibitem{chattopadhyay2017counting}
P.~Chattopadhyay, R.~Vedantam, R.~R. Selvaraju, D.~Batra, and D.~Parikh.
\newblock Counting everyday objects in everyday scenes.
\newblock In {\em CVPR}, pages 4428--4437, 2017.

\bibitem{chen2013neil}
X.~Chen, A.~Shrivastava, and A.~Gupta.
\newblock Neil: Extracting visual knowledge from web data.
\newblock In {\em Proceedings of the IEEE International Conference on Computer
  Vision}, pages 1409--1416, 2013.

\bibitem{cheron2018flexible}
G.~Ch{\'e}ron, J.-B. Alayrac, I.~Laptev, and C.~Schmid.
\newblock A flexible model for training action localization with varying levels
  of supervision.
\newblock {\em arXiv preprint arXiv:1806.11328}, 2018.

\bibitem{10}
R.~G. Cinbis, J.~Verbeek, and C.~Schmid.
\newblock Weakly supervised object localization with multi-fold multiple
  instance learning.
\newblock {\em TPAMI}, 39(1):189--203, 2017.

\bibitem{collomosse2017sketching}
J.~Collomosse, T.~Bui, M.~J. Wilber, C.~Fang, and H.~Jin.
\newblock Sketching with style: Visual search with sketches and aesthetic
  context.
\newblock In {\em ICCV}, pages 2679--2687, 2017.

\bibitem{r-fcn}
J.~Dai, Y.~Li, K.~He, and J.~Sun.
\newblock {R-FCN:} object detection via region-based fully convolutional
  networks.
\newblock {\em CoRR}, abs/1605.06409, 2016.

\bibitem{du2017fused}
X.~Du, M.~El-Khamy, J.~Lee, and L.~Davis.
\newblock Fused dnn: A deep neural network fusion approach to fast and robust
  pedestrian detection.
\newblock In {\em Applications of Computer Vision (WACV), 2017 IEEE Winter
  Conference on}, pages 953--961. IEEE, 2017.

\bibitem{pascal}
M.~Everingham, L.~Van~Gool, C.~K. Williams, J.~Winn, and A.~Zisserman.
\newblock The pascal visual object classes (voc) challenge.
\newblock {\em IJCV}, 88(2):303--338, 2010.

\bibitem{fast-rcnn}
R.~Girshick.
\newblock Fast r-cnn.
\newblock In {\em ICCV}, pages 1440--1448, 2015.

\bibitem{rcnn}
R.~B. Girshick, J.~Donahue, T.~Darrell, and J.~Malik.
\newblock Rich feature hierarchies for accurate object detection and semantic
  segmentation.
\newblock {\em CoRR}, abs/1311.2524, 2013.

\bibitem{15}
R.~Gokberk~Cinbis, J.~Verbeek, and C.~Schmid.
\newblock Multi-fold mil training for weakly supervised object localization.
\newblock In {\em CVPR}, pages 2409--2416, 2014.

\bibitem{hasan2015context}
M.~Hasan and A.~K. Roy-Chowdhury.
\newblock Context aware active learning of activity recognition models.
\newblock In {\em Proceedings of the IEEE International Conference on Computer
  Vision}, pages 4543--4551, 2015.

\bibitem{mask-rcnn}
K.~He, G.~Gkioxari, P.~Doll{\'a}r, and R.~Girshick.
\newblock Mask r-cnn.
\newblock {\em arXiv preprint arXiv:1703.06870}, 2017.

\bibitem{heilbron2018annotate}
F.~C. Heilbron, J.-Y. Lee, H.~Jin, and B.~Ghanem.
\newblock What do i annotate next? an empirical study of active learning for
  action localization.
\newblock In {\em ECCV}, 2018.

\bibitem{18}
J.~Hoffman, D.~Pathak, T.~Darrell, and K.~Saenko.
\newblock Detector discovery in the wild: Joint multiple instance and
  representation learning.
\newblock In {\em CVPR}, pages 2883--2891, 2015.

\bibitem{hu2017finding}
P.~Hu and D.~Ramanan.
\newblock Finding tiny faces.
\newblock In {\em CVPR}, pages 1522--1530. IEEE, 2017.

\bibitem{22}
Z.~Jie, Y.~Wei, X.~Jin, J.~Feng, and W.~Liu.
\newblock Deep self-taught learning for weakly supervised object localization.
\newblock {\em arXiv preprint arXiv:1704.05188}, 2017.

\bibitem{kang2018crowd}
D.~Kang and A.~Chan.
\newblock Crowd counting by adaptively fusing predictions from an image
  pyramid.
\newblock {\em arXiv preprint arXiv:1805.06115}, 2018.

\bibitem{11}
V.~Kantorov, M.~Oquab, M.~Cho, and I.~Laptev.
\newblock Contextlocnet: Context-aware deep network models for weakly
  supervised localization.
\newblock In {\em ECCV}, pages 350--365. Springer, 2016.

\bibitem{kao2018localization}
C.-C. Kao, T.-Y. Lee, P.~Sen, and M.-Y. Liu.
\newblock Localization-aware active learning for object detection.
\newblock {\em arXiv preprint arXiv:1801.05124}, 2018.

\bibitem{kapoor2007active}
A.~Kapoor, K.~Grauman, R.~Urtasun, and T.~Darrell.
\newblock Active learning with gaussian processes for object categorization.
\newblock In {\em ICCV}, pages 1--8. IEEE, 2007.

\bibitem{konyushkova2015introducing}
K.~Konyushkova, R.~Sznitman, and P.~Fua.
\newblock Introducing geometry in active learning for image segmentation.
\newblock In {\em Proceedings of the IEEE International Conference on Computer
  Vision}, pages 2974--2982, 2015.

\bibitem{konyushkova2017lal}
K.~Konyushkova, R.~Sznitman, and P.~Fua.
\newblock Learning active learning from data.
\newblock In {\em Advances in Neural Information Processing Systems}, pages
  4225--4235, 2017.

\bibitem{konyushkova2018learning}
K.~Konyushkova, J.~Uijlings, C.~H. Lampert, and V.~Ferrari.
\newblock Learning intelligent dialogs for bounding box annotation.
\newblock In {\em CVPR}, 2018.

\bibitem{laine2016temporal}
S.~Laine and T.~Aila.
\newblock Temporal ensembling for semi-supervised learning.
\newblock {\em arXiv preprint arXiv:1610.02242}, 2016.

\bibitem{USSampling}
D.~D. Lewis and J.~Catlett.
\newblock Heterogeneous uncertainty sampling for supervised learning.
\newblock In {\em Machine Learning Proceedings 1994}, pages 148--156. Elsevier,
  1994.

\bibitem{17}
D.~Li, J.-B. Huang, Y.~Li, S.~Wang, and M.-H. Yang.
\newblock Weakly supervised object localization with progressive domain
  adaptation.
\newblock In {\em CVPR}, pages 3512--3520, 2016.

\bibitem{li2010optimol}
L.-J. Li and L.~Fei-Fei.
\newblock Optimol: automatic online picture collection via incremental model
  learning.
\newblock {\em International journal of computer vision}, 88(2):147--168, 2010.

\bibitem{fpn}
T.~Lin, P.~Doll{\'{a}}r, R.~B. Girshick, K.~He, B.~Hariharan, and S.~J.
  Belongie.
\newblock Featuref pyramid networks for object detection.
\newblock {\em CoRR}, abs/1612.03144, 2016.

\bibitem{coco}
T.-Y. Lin, M.~Maire, S.~Belongie, J.~Hays, P.~Perona, D.~Ramanan,
  P.~Doll{\'a}r, and C.~L. Zitnick.
\newblock Microsoft coco: Common objects in context.
\newblock In {\em ECCV}, pages 740--755. Springer, 2014.

\bibitem{ActiveHumanPose}
B.~Liu and V.~Ferrari.
\newblock Active learning for human pose estimation.
\newblock In {\em Proceedings of the IEEE International Conference on Computer
  Vision}, pages 4363--4372, 2017.

\bibitem{liu2016ssd}
W.~Liu, D.~Anguelov, D.~Erhan, C.~Szegedy, S.~Reed, C.-Y. Fu, and A.~C. Berg.
\newblock Ssd: Single shot multibox detector.
\newblock In {\em ECCV}, pages 21--37. Springer, 2016.

\bibitem{misra2017learning}
I.~Misra, R.~Girshick, R.~Fergus, M.~Hebert, A.~Gupta, and L.~van~der Maaten.
\newblock Learning by asking questions.
\newblock {\em arXiv preprint arXiv:1712.01238}, 2017.

\bibitem{mu2018towards}
C.~Mu, J.~Zhao, G.~Yang, J.~Zhang, and Z.~Yan.
\newblock Towards practical visual search engine within elasticsearch.
\newblock {\em arXiv preprint arXiv:1806.08896}, 2018.

\bibitem{papadopoulos2016we}
D.~P. Papadopoulos, J.~R. Uijlings, F.~Keller, and V.~Ferrari.
\newblock We don't need no bounding-boxes: Training object class detectors
  using only human verification.
\newblock In {\em Proceedings of the IEEE Conference on Computer Vision and
  Pattern Recognition}, pages 854--863, 2016.

\bibitem{papadopoulos2017extreme}
D.~P. Papadopoulos, J.~R. Uijlings, F.~Keller, and V.~Ferrari.
\newblock Extreme clicking for efficient object annotation.
\newblock In {\em ICCV}, pages 4940--4949. IEEE, 2017.

\bibitem{qi2008two}
G.-J. Qi, X.-S. Hua, Y.~Rui, J.~Tang, and H.-J. Zhang.
\newblock Two-dimensional active learning for image classification.
\newblock In {\em CVPR}, pages 1--8. IEEE, 2008.

\bibitem{radosavovic2017data}
I.~Radosavovic, P.~Doll{\'a}r, R.~Girshick, G.~Gkioxari, and K.~He.
\newblock Data distillation: Towards omni-supervised learning.
\newblock {\em arXiv preprint arXiv:1712.04440}, 2017.

\bibitem{hyperface}
R.~Ranjan, V.~M. Patel, and R.~Chellappa.
\newblock Hyperface: A deep multi-task learning framework for face detection,
  landmark localization, pose estimation, and gender recognition.
\newblock {\em IEEE Transactions on Pattern Analysis and Machine Intelligence},
  pages 1--1, 2018.

\bibitem{yolov1}
J.~Redmon, S.~Divvala, R.~Girshick, and A.~Farhadi.
\newblock You only look once: Unified, real-time object detection.
\newblock In {\em CVPR}, pages 779--788, 2016.

\bibitem{redmon2016yolo9000}
J.~Redmon and A.~Farhadi.
\newblock Yolo9000: Better, faster, stronger.
\newblock {\em arXiv preprint arXiv:1612.08242}, 2016.

\bibitem{renNIPS15fasterrcnn}
S.~Ren, K.~He, R.~Girshick, and J.~Sun.
\newblock Faster {R-CNN}: Towards real-time object detection with region
  proposal networks.
\newblock In {\em Advances in Neural Information Processing Systems ({NIPS})},
  2015.

\bibitem{19}
M.~Rochan and Y.~Wang.
\newblock Weakly supervised localization of novel objects using appearance
  transfer.
\newblock In {\em CVPR}, pages 4315--4324, 2015.

\bibitem{rosenberg2005semi}
C.~Rosenberg, M.~Hebert, and H.~Schneiderman.
\newblock Semi-supervised self-training of object detection models.
\newblock In {\em WACV/MOTION}, pages 29--36, 2005.

\bibitem{imagenet}
O.~Russakovsky, J.~Deng, H.~Su, J.~Krause, S.~Satheesh, S.~Ma, Z.~Huang,
  A.~Karpathy, A.~Khosla, M.~Bernstein, et~al.
\newblock Imagenet large scale visual recognition challenge.
\newblock {\em IJCV}, 115(3):211--252, 2015.

\bibitem{16}
H.~O. Song, Y.~J. Lee, S.~Jegelka, and T.~Darrell.
\newblock Weakly-supervised discovery of visual pattern configurations.
\newblock In {\em Advances in Neural Information Processing Systems ({NIPS})},
  pages 1637--1645, 2014.

\bibitem{AAAIW125350}
H.~Su, J.~Deng, and L.~Fei-Fei.
\newblock Crowdsourcing annotations for visual object detection.
\newblock In {\em Workshops at the Twenty-Sixth AAAI Conference on Artificial
  Intelligence}, volume~1, 2012.

\bibitem{thorpe1996speed}
S.~Thorpe, D.~Fize, and C.~Marlot.
\newblock Speed of processing in the human visual system.
\newblock {\em nature}, 381(6582):520, 1996.

\bibitem{20}
C.~Wang, W.~Ren, K.~Huang, and T.~Tan.
\newblock Weakly supervised object localization with latent category learning.
\newblock In {\em ECCV}, pages 431--445. Springer, 2014.

\bibitem{Wang2014ANA}
D.~Wang and Y.~Shang.
\newblock A new active labeling method for deep learning.
\newblock {\em 2014 International Joint Conference on Neural Networks (IJCNN)},
  pages 112--119, 2014.

\bibitem{wang2018pcn}
S.~Wang, J.~Cheng, H.~Liu, and M.~Tang.
\newblock Pcn: Part and context information for pedestrian detection with cnns.
\newblock {\em arXiv preprint arXiv:1804.04483}, 2018.

\bibitem{watkins1992q}
C.~J. Watkins and P.~Dayan.
\newblock Q-learning.
\newblock {\em Machine learning}, 8(3-4):279--292, 1992.

\bibitem{jjfaster2rcnn}
J.~Yang, J.~Lu, D.~Batra, and D.~Parikh.
\newblock A faster pytorch implementation of faster r-cnn.
\newblock {\em https://github.com/jwyang/faster-rcnn.pytorch}, 2017.

\bibitem{zhang2018w2f}
Y.~Zhang, Y.~Bai, M.~Ding, Y.~Li, and B.~Ghanem.
\newblock W2f: A weakly-supervised to fully-supervised framework for object
  detection.
\newblock In {\em Proceedings of the IEEE Conference on Computer Vision and
  Pattern Recognition}, pages 928--936, 2018.

\bibitem{zhu2017generative}
J.-J. Zhu and J.~Bento.
\newblock Generative adversarial active learning.
\newblock {\em arXiv preprint arXiv:1702.07956}, 2017.

\bibitem{14}
B.~Zhuang, L.~Liu, Y.~Li, C.~Shen, and I.~Reid.
\newblock Attend in groups: a weakly-supervised deep learning framework for
  learning from web data.
\newblock {\em arXiv preprint arXiv:1611.09960}, 2016.

\end{thebibliography}
}
\end{appendices}

\end{document}